\documentclass{article}

\usepackage[preprint, nonanonymous]{neurips_2026}
\usepackage{paralist}

\usepackage[utf8]{inputenc} %
\usepackage[T1]{fontenc}    %
\usepackage{hyperref}       %
\usepackage{url}            %
\usepackage{booktabs}       %
\usepackage{amsfonts}       %
\usepackage{nicefrac}       %
\usepackage{microtype}      %
\usepackage[table]{xcolor}         %

\usepackage{graphicx}
\usepackage{enumitem}
\usepackage{caption}
\usepackage{subcaption} 
\usepackage{amsmath}
\usepackage{tabularx}
\usepackage{makecell} %
\usepackage{multirow} 
\usepackage{tcolorbox}
\tcbuselibrary{skins, breakable}

\usepackage{xcolor} 
\definecolor{red}{HTML}{FF8080}
\definecolor{green}{HTML}{72C3A3}
\colorlet{worse}{red!40}
\colorlet{better}{green!40}

\usepackage{pifont}
\newcommand{\cmark}{\ding{51}}
\newcommand{\xmark}{\ding{55}}

\newcommand{\ourdata}{{CAREBench}}

\title{CAREBench: Evaluating LLMs' Emotion Understanding by Assessing Cognitive Appraisal Reasoning}

\author{%
  Zhaoyue Sun \\
  Department of Informatics\\
  King's College London\\
  \texttt{zhaoyue.1.sun@kcl.ac.uk} \\
  \And
  Hainiu Xu \\
  Department of Informatics\\
  King's College London\\
  \texttt{hainiu.xu@kcl.ac.uk} \\
  \And
  Andero Uusberg \\
  Institute of Psychology\\
  University of Tartu\\
  \texttt{andero.uusberg@ut.ee} \\
  \And
  James J. Gross \\
  Department of Psychology\\
  Stanford University\\
  \texttt{gross@stanford.edu} \\
  \And
  Petr Slovak\thanks{Corresponding authors: Petr Slovak and Yulan He.} \\
  Department of Informatics\\
  King's College London\\
  \texttt{petr.slovak@kcl.ac.uk} \\
  \And
  Yulan He\footnotemark[1] \\
  Department of Informatics\\
  King's College London\\
  The Alan Turing Institute\\
  \texttt{yulan.he@kcl.ac.uk} \\
}

\begin{document}

\maketitle

\begin{abstract}
  
Emotion understanding is a core capability for LLMs to interact effectively with humans, yet existing evaluation paradigms rely on discrete emotion label prediction and fail to capture the cognitive processes underlying emotion generation. Grounded in appraisal theory, we introduce \ourdata, the first benchmark with complete inferential chain annotations from both first- and third-person perspectives on real-world narratives, spanning appraisal reasoning, appraisal ratings, and multi-label emotion annotation. We propose a process-level evaluation framework and conduct systematic experiments across six LLMs organized around four research questions. We find that stronger models match or surpass human observers on certain tasks, yet fall short on appraisal reasoning and positive emotion recognition; performance across chain steps and sensitivity to appraisal interventions exhibit dissociations across models; and current models have not internalized the mechanisms needed to capture human subjective heterogeneity. These findings suggest that downstream emotion prediction metrics may overestimate LLMs' true emotion understanding, and \ourdata~provides a foundation for more diagnostically informative evaluation of LLMs' affective cognitive capabilities.
\end{abstract}

\section{Introduction}

Emotion understanding is a fundamental capability for AI systems built on Large Language Models (LLMs) deployed in affective computing, human-AI interaction, and mental health support. However, evaluating this capability remains challenging due to the complexity of how emotions are elicited, experienced, and expressed in humans. Most prior work has assessed emotion understanding through discrete emotion label prediction. However, this fails to capture the richness of real-world emotional phenomena \cite{bonard2024improving}. To address this limitation, recent work has turned to psychological theories of emotion. Among these, appraisal theory posits that emotions arise from individuals’ subjective evaluations of events along multiple dimensions — such as goal relevance and coping potential — providing a structured and interpretable representation for evaluating emotion understanding \cite{scherer1999appraisal}. Under this perspective, recent studies have introduced appraisal-grounded datasets to examine whether LLMs exhibit cognitive emotion understanding capabilities beyond label prediction \cite{troiano2023dimensional, liu2025cape, bhattacharyya2026do}.

However, existing appraisal-grounded datasets exhibit three key limitations. First, they predominantly treat appraisal dimension ratings as the sole representation of cognitive appraisal. However, much of  appraisal involves  inferences over mental states, such as beliefs and desires, while appraisal ratings reflect only its outcomes. Without explicitly modeling this inference process, it remains unclear whether a model truly understands the experiencer’s mental states or merely learns a superficial mapping from situations to ratings. Second, most datasets rely on single emotion category labels, whereas real-world emotional experiences are often mixed and involve multiple co-occurring emotions; thus, a single label fails to capture this richness. Third, the majority of datasets collect only third-person annotations, making it difficult to examine how perspective differences between experiencers and observers systematically influence emotion judgments. 
While the Crowd-Envent dataset \cite{troiano2023dimensional} includes both first- and third-person perspectives, it contains very short situation descriptions (on average 20 words) and provides only appraisal dimension ratings. As a result, it cannot capture variation in annotators’ underlying appraisal inference processes.

\begin{figure} [th]
    \centering
    \includegraphics[width=0.8\linewidth]{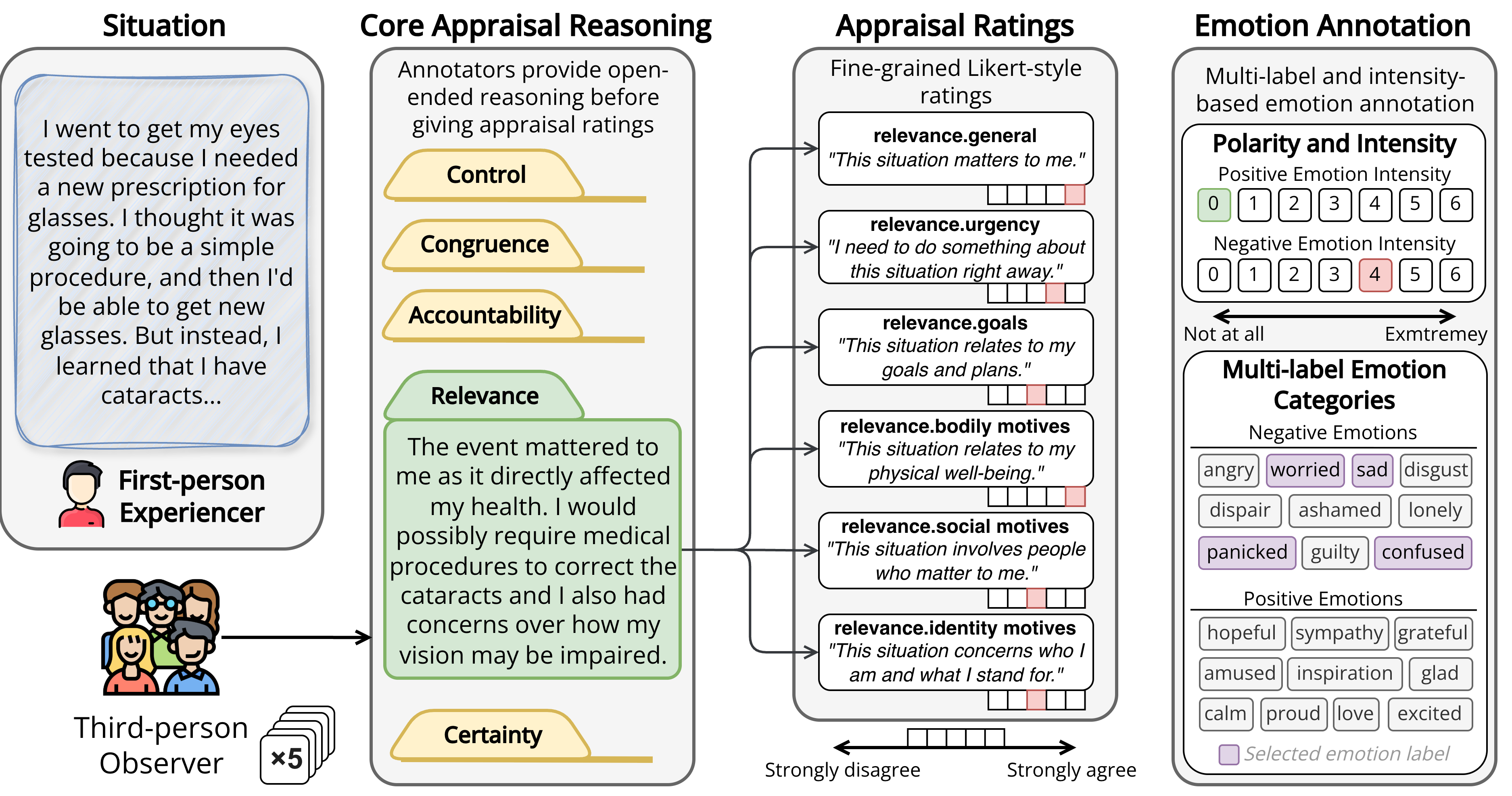}
    \caption{Overview of the data annotation pipeline.}
    \label{fig:figure1}
\end{figure}

To address these gaps, we constructed \ourdata, a benchmark designed to evaluate Cognitive Appraisal Reasoning in Emotion understanding, with \textbf{three key features} (Figure~\ref{fig:figure1}). First, we explicitly annotated \textbf{the reasoning process of appraisals}. Specifically, we recruited annotators to provide real-life emotional narratives (95 words on average), followed by three annotation layers collected in sequence: open-ended reasoning texts targeting five core appraisal dimensions, hierarchical appraisal ratings across 22 sub-dimensions, and emotion annotations.
This design enables the construction of a situation → reasoning → appraisal → emotion inferential chain, making it possible to observe mental-state layer of appraisal, analyze its influence on fine-grained appraisal ratings, and evaluate LLMs' emotion understanding correctness at the process level. Probing mental states before appraisal ratings also scaffolds the annotation process itself, ensuring that ratings are grounded in explicit reasoning rather than intuitive impression. 
To our knowledge, no existing dataset elicits mental states as a precursor to appraisal annotation, in either first-person or third-person settings.
Second, for emotion annotation, instead of representing emotion as a single discrete label, we allow \textbf{multi-label emotion annotation with intensity}. %
This supports many-to-many analysis of appraisal–emotion mappings and more faithful modeling of complex real-world emotional responses. %
In addition, we provide \textbf{full inferential-chain annotations from both the first-person experiencer and multiple third-person observers}, enabling direct comparison between LLMs and both perspectives, as well as localization of subjective heterogeneity along the whole process. Annotator demographics and personality traits are also collected to support subjectivity analysis.

Based on this dataset, we conducted systematic evaluations of six LLMs organized around four progressively structured research questions: how well models perform each inferential chain step, whether injecting appraisal reasoning helps with downstream prediction, whether models exhibit directional sensitivity to structured appraisal interventions, and whether they can capture the distributional heterogeneity of human appraisal judgments. Our key findings are: (1) although stronger models achieve competitive downstream performance, appraisal reasoning capabilities remain limited and positive emotion identification is universally challenging; (2) capabilities across chain steps are dissociable among models, with upstream advantages not consistently transferring downstream; (3) emotion prediction performance does not fully align with sensitivity to appraisal interventions, suggesting some models rely on associative rather than appraisal-based reasoning; and (4) current models have not internalized the mechanisms needed to capture subjective heterogeneity.

The main contributions of this work are as follows:
\begin{itemize}[leftmargin=*,noitemsep]
    \item \textbf{Dataset.} We introduce \ourdata, a new benchmark for evaluating LLMs' emotion understanding, supplying data absent in existing datasets on explicit appraisal reasoning, multi-emotion co-occurrence, and dual-perspective emotion understanding.
    \item \textbf{Evaluation framework.} We propose a process-level evaluation framework covering the full situation $\rightarrow$ reasoning $\rightarrow$ appraisal $\rightarrow$ emotion inferential chain, moving beyond the single-step, single-perspective evaluation paradigms in existing work. This framework exposes the blind spots of relying solely on downstream emotion prediction performance to assess LLMs' emotion understanding, providing methodological support for more reliable evaluation of LLMs' affective cognitive capabilities.
    \item \textbf{Empirical findings.} Through systematic evaluation of six LLMs, we provide an empirical basis for comprehensively diagnosing current LLMs' emotion understanding capabilities, revealing the insufficiency of existing evaluation paradigms in distinguishing surface-level prediction performance from genuine appraisal understanding, with direct implications for building more cognitively interpretable affective models in the future.
\end{itemize}

\section{The CAREBench Dataset}

\subsection{Data Collection}
\label{sec:data}

Process models of appraisal theory characterize emotion generation as a multi-stage sequential process, spanning a complete cognitive chain from situation construal through appraisal to emotional response~\cite{scherer2001appraisal, marsella2009ema}. Grounded in this theoretical perspective, we collected annotations covering an inferential chain $s \rightarrow r \rightarrow \mathbf{a} \rightarrow e$, where $s$ denotes the situational narrative, $r$ the open-ended appraisal reasoning text, $\mathbf{a}$ the appraisal rating vector, $e$ the emotion annotation.
In collaboration with psychology experts, we designed a two-phase data collection protocol, with participants recruited via the Prolific platform.
In the first phase, participants provided first-person emotional narratives and completed self-annotation; in the second phase, independent third-party annotators read these narratives and completed annotation from an observer's perspective. %

\paragraph{Phase 1: First-Person Narratives and Annotation}
In Phase 1, each participant was asked to recall and describe a recent emotionally significant event. To ensure richness of content while reducing participant burden, we adopted the micro-narratives framework \cite{skeggs2025micro}, using an LLM (GPT-4o) to assist participants in composing their narratives. Specifically, participants engaged in a structured conversation with an AI chatbot, answering a series of guided questions; the LLM then assembled the resulting text by combining participants' responses in their own words with minimal reformulation, which participants reviewed and corrected as needed. Participants subsequently completed a three-layer annotation task: (1) appraisal reasoning  ($r$): open-ended responses to questions targeting five core appraisal dimensions (relevance, certainty, congruence, control, and accountability), also completed via the micro-narratives framework; (2) appraisal ratings ($\mathbf{a} \in \mathbb{R}^{22}$): 5-point Likert-scale ratings on 22 appraisal items derived through hierarchical expansion of the five core dimensions; (3) emotion annotation ($e$): participants rated positive and negative emotion intensity $e_i^+$ and $e_i^-$ on a 7-point Likert scale, and selected positive and negative emotion labels $e_l^+$ and $e_l^-$ from a predefined label set $\mathcal{L}$, with multiple labels and co-occurring positive and negative emotions permitted. In addition, each participant completed the Big Five Inventory-10 (BFI-10) \cite{rammstedt2007measuring}, which, together with demographic information provided by Prolific, served to characterize each annotator's persona.

\paragraph{Phase 2: Third-Person Annotation}
In Phase 2, we aimed to collect multiple third-person annotations per narrative to capture cross-individual variation in emotional interpretation; five annotations were finally collected for a subset of 500 first-person narratives owing to cost constraints. For each narrative, annotators read only the narrative text, without access to the first-person appraisal responses or ratings, and independently completed the same three-layer annotation task along with BFI-10 and demographic information. %

The total expenditure amounted to approximately £9,000, including £4.20 per first-person annotation, £1.68 per third-person annotation (both including platform fees), and pilot study and API costs. Full annotation schema and other details of the data collection procedure are provided in Appendix~\ref{app:data-collection}.

\subsection{Data Statistics}

\begin{figure}[t]
\centering

\begin{minipage}[t]{0.49\textwidth}
  \centering

  \captionof{table}{%
    Dataset statistics. `Avg Len.' is measured in tokens; `+ Apr' includes core appraisal reasoning; `Avg lab.' denotes the averaged number of emotion labels. `Pos./Neg. Samples' indicates samples with positive/negative emotions.
  }
  \label{tab:data_statistics}
  \vspace{4pt}
  {\setlength{\tabcolsep}{3pt}
  \scriptsize
  \begin{tabular}{lcccccccc}
    \toprule
    & Ann.
    & \multicolumn{2}{c}{Avg Len.\ (tok.)}
    & \multicolumn{2}{c}{Pos.\ Samples}
    & \multicolumn{2}{c}{Neg.\ Samples} \\
    \cmidrule(lr){3-4}\cmidrule(lr){5-6}\cmidrule(lr){7-8}
    & Count & Story & +Apr. & Count & Avg lab. & Count & Avg lab. \\
    \midrule
    1st & 1{,}000 & 95.1 & 168.5 & 456   & 4.05 & 771   & 3.43 \\
    3rd & 2{,}500 & ---  & 160.6 & 1{,}232 & 3.38 & 1{,}932 & 3.17 \\
    \bottomrule
  \end{tabular}}

  \vspace{10pt}

  \includegraphics[width=\linewidth]{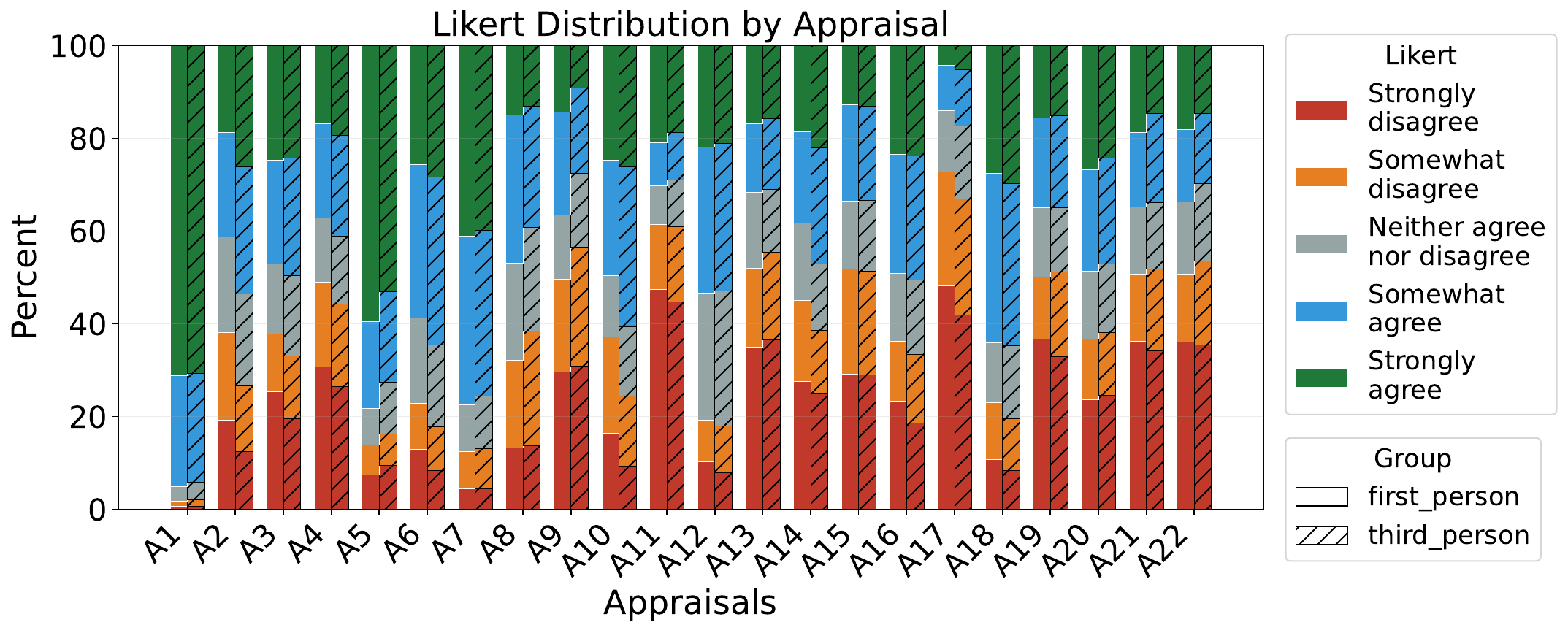}
  \captionof{figure}{%
    Likert-scale response distributions across 22 appraisal dimensions from Table~\ref{tab:appraisal-items}.
  }
  \label{fig:appraisal_likert}

\end{minipage}
\hfill
\begin{minipage}[t]{0.49\textwidth}
  \centering

  \captionof{figure}{%
    Emotion annotation statistics.
    (a)~Intensity score distributions.
    (b)~Number of emotion labels per sample.
    (c)~Distribution of positive emotion categories.
    (d)~Distribution of negative emotion categories.
  }
  \label{fig:emotion_statistics}
  \vspace{4pt}
  \includegraphics[width=\linewidth]{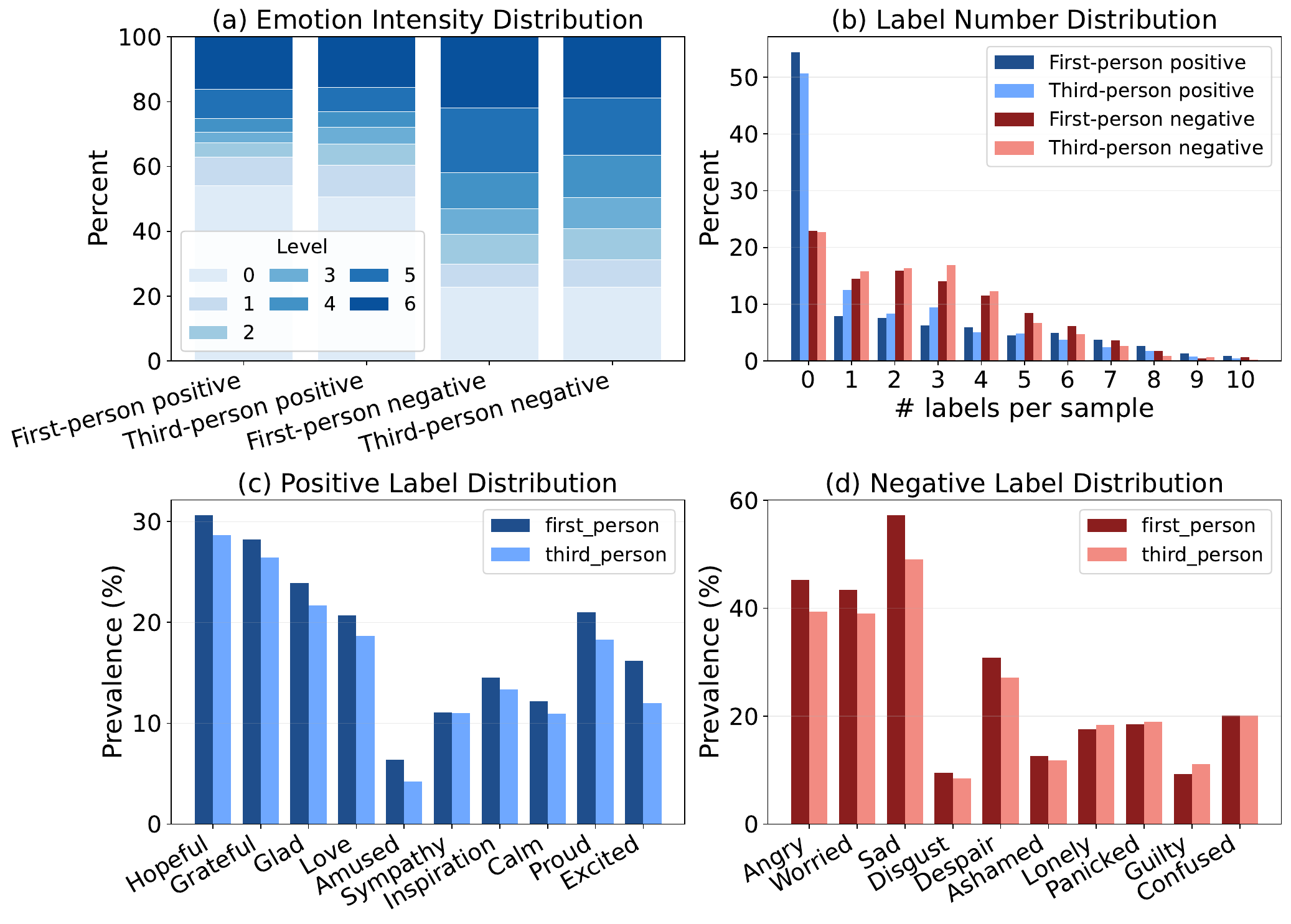}

\end{minipage}

\end{figure}

The dataset comprises 1,000 first-person annotated samples, 500 of which are additionally annotated from the third-person perspective by 5 independent annotators each, yielding 2,500 third-person annotation records in total (Table~\ref{tab:data_statistics}). The average story length is 95.1 tokens; with core appraisal reasoning text included, the average reaches approximately 160 tokens, with third-person annotators producing slightly shorter inference texts than their first-person counterparts.

In terms of appraisal rating distributions (Figure~\ref{fig:appraisal_likert}), the two annotator groups exhibit broadly similar patterns, though notable polarity shifts emerge on certain dimensions. For instance, third-person annotators tend to overestimate event urgency on A2 (\emph{relevance.urgency}). At the dataset level, most appraisal dimensions show relatively balanced rating distributions, while a few skew markedly toward one pole: A1 (\emph{relevance.general}) is consistently high, indicating that most situations are perceived as highly relevant to the protagonist, whereas A17 (\emph{control.vicarious}) is consistently low, reflecting a widespread perception that others are unlikely to intervene in the current situation.

Regarding emotion annotations (Table~\ref{tab:data_statistics} \& Figure~\ref{fig:emotion_statistics}), negative-emotion samples outnumber positive ones by approximately 1.6-fold, reflecting the natural skew of emotional narratives. Label co-occurrence is prevalent, with most samples receiving 1–4 labels; positive samples carry more labels on average than negative ones, suggesting that positive contexts more readily elicit mixed emotions. Third-person observers tend to assign fewer emotion labels overall, and their rated intensity for negative emotions is slightly lower than that of first-person experiencers, while positive emotion intensity shows a more dispersed distribution. High-frequency emotion categories are highly consistent across perspectives, yet third-person prevalence is lower than first-person for the vast majority of categories, with only a handful of labels (\emph{lonely}, \emph{panicked}, \emph{guilty}, \emph{confused}) being relatively more salient from the third-person perspective.

\section{Experimental Setup}
\label{sec:exp}
\subsection{Models}

We evaluate six models spanning three categories. For closed-source models, we include GPT-5.2 (GPT) \cite{openai2025gpt} and Claude-Sonnet-4.6 (Claude) \cite{anthropic2026claude}. For open-source general-purpose models, we include Llama 3.1-8B-Instruct (Llama) \cite{grattafiori2024llama} and Qwen 3.5-9B (Qwen) \cite{qwen35blog}. To examine whether task-specific training on psychology- or emotion-related data confers advantages on appraisal-grounded emotion understanding, we include PsyLLM \cite{hu2025beyond}, fine-tuned from Qwen 3-8B on synthetic mental health counseling dialogues derived from Reddit posts, and EmoLLaMa-chat-13B (EmoLlama) \cite{liu2024emollms}, an instruction-tuned model built on LLaMA2-chat-13B using affective analysis data. More reproducibility details are provided in Appendix~\ref{app:reprod}.

\subsection{Evaluation Metrics}

Our evaluation framework mirrors the structure of the emotion inferential chain, with metrics defined at each step from situation construal to emotion output, supplemented by measures of subjectivity modeling. Specifically, we evaluate \begin{inparaenum}[1)]
    \item \textbf{Core appraisal reasoning}: since this task requires open-ended text generation, we report BLEU, ROUGE-1, ROUGE-2, ROUGE-L, and BERTScore;
    \item \textbf{Appraisal prediction}: we compute normalized root mean square error (RMSE) for each appraisal item, where ratings are on a 5-point Likert scale;
    \item \textbf{Emotion intensity prediction}: we likewise use normalized RMSE, where intensity ratings are on a 7-point Likert scale;
    \item \textbf{Emotion label prediction}: we report example-F1 (per-instance F1 averaged across the dataset), micro-F1 (micro-averaged across labels), and macro-F1 (macro-averaged across labels);
    \item \textbf{Subjectivity modeling}: since appraisal maps emotion understanding onto a measurable low-dimensional space, providing a convenient medium for assessing subjective heterogeneity, all metrics here are defined over appraisal dimensions. Following Zhou et al.~\cite{zhou2025modeling}, we use $\mu$-MAE, $\sigma$-MAE, and Wasserstein distance to measure the discrepancy between the model-simulated and human annotator appraisal distributions. We additionally introduce Spearman correlation and precision@$k$ to assess whether models can identify which appraisal dimensions are more prone to subjective disagreement.
\end{inparaenum}

\subsection{Research Questions and Experimental Design}

\textbf{RQ1: How well do LLMs perform each step of the inference chain?}

We evaluate LLM prediction performance at each node of the inferential chain under zero-shot settings, covering three tasks: inferring appraisal reasoning text ($s \rightarrow r$), predicting appraisal ratings ($s \rightarrow \mathbf{a}$), and predicting emotion outputs ($s \rightarrow e$). In each task, models are prompted to put themselves in the shoes of the experiencer, as human third-person observers are instructed to do, and generate predictions accordingly. Model outputs are then compared against human third-person annotations on the same tasks, identifying systematic divergences between model and human observer performance at each step.

\textbf{RQ2: Does the appraisal reasoning process help improve appraisal prediction and emotion recognition?}

Appraisal theory posits that the relationship between situational information and emotional outcomes is mediated by structured cognitive appraisals, which has been widely validated in studies on human participants~\cite{smith1985patterns, roseman1991appraisal}. The role of appraisal in LLM-based emotion recognition has also been preliminarily explored in prior work~\cite{troiano2023dimensional, debnath2025appraisal, tak2025mechanistic}. Given this, we examine whether appraisal reasoning $r$, open-ended text assumed to be more accessible to LLMs, could improve LLMs' performance on appraisal prediction and emotion recognition. Specifically, we take situational narrative, $s$, as the baseline condition and design four augmented input conditions: $s+r$, $s+\mathbf{a}$, $s+r+\mathbf{a}$, and $s+\hat{r}$, where $\hat{r}$ denotes model-generated appraisal reasoning. We compare $\Delta P(\mathbf{a})$ and $\Delta P(e)$ across the four conditions, where $P(\cdot)$ denotes the corresponding evaluation metric and $\Delta$ denotes the change relative to the baseline $s$, assessing whether and how reasoning text improve downstream predictions relative to $\mathbf{a}$ and $\hat{r}$. Further, for each appraisal dimension we inject dimension-specific reasoning text $r_k$ into $s$ and examine its differential effects on various dimensions of $\mathbf{a}$ and categories of $e$, testing whether $r_k$ yields targeted improvements on the certain appraisals and emotions.

\textbf{RQ3: Are emotion predictions in LLMs sensitive to structured appraisal interventions?} %

RQ3 examines  whether LLM predictions are sensitive to controlled interventions on appraisal reasoning text. For each appraisal dimension, we select samples in which the first-person rating and third-person rating fall on opposite sides of the scale midpoint, and substitute the input reasoning text from the first-person $r^{(1)}$ to the third-person $r^{(3)}$, examining model prediction changes along two pathways: $r\rightarrow\mathbf{a}$ and $r\rightarrow e$. For $r\rightarrow\mathbf{a}$, we compute the Pearson correlation between $\Delta\hat{\mathbf{a}} = f(s,r^{(3)}) - f(s,r^{(1)})$ and $\Delta\mathbf{a} = \mathbf{a}^{(3)} - \mathbf{a}^{(1)}$. For $r\rightarrow e$, we compute the correlation between $\Delta\hat{e}_i^\pm$ and $\Delta e_i^\pm = e_i^{\pm(3)} - e_i^{\pm(1)}$, and between $\Delta\hat{p}_l$ and $\Delta p_l$, where $\Delta p_l$ denotes the change in emotion label occurrence rate across qualifying samples per dimension. The resulting patterns are then compared against appraisal-emotion mappings predicted by appraisal theory. When multiple third-person annotators qualify for the same scenario-dimension pair, all change quantities are averaged before computing the correlation. While this does not prove causal reasoning, consistent alignment indicates sensitivity to appraisal-relevant information beyond label associations.

\textbf{RQ4: Do LLMs capture the distributional heterogeneity of human appraisal judgments?}

The preceding RQs treat first-person annotations as the target for model inference, but emotion understanding is inherently heterogeneous — different observers may systematically differ in their appraisal judgments of the same situation. RQ4 examines whether LLMs can approximate the distributional heterogeneity of human appraisal judgments. Let $\mathcal{D}^{(3)}$ denote the distribution of third-person appraisal ratings for a given scenario $s$. We obtain LLM prediction distributions $\hat{\mathcal{D}}_c$ under three sampling conditions: repeated sampling ($c_\text{base}$), sampling conditioned on injected persona information ($c_\text{persona}$), and sampling conditioned on injected core appraisal reasoning ($c_\text{reason}$), where persona information consists of demographic attributes and Big Five personality traits converted into a biographical profile via a template. First, we measure the match between $\hat{\mathcal{D}}_c$ and $\mathcal{D}^{(3)}$ using $\mu$-MAE, $\sigma$-MAE, and Wasserstein distance, comparing distribution fitting quality across the three conditions. Second, we assess whether LLMs can identify variation in inter-annotator disagreement across dimensions and scenarios, measured by the rank correlation between $\text{Var}(\hat{\mathcal{D}}_c)$ and $\text{Var}(\mathcal{D}^{(3)})$ across dimensions.

\section{Results and Analysis}
\label{sec:results}

\textbf{RQ1: How well do LLMs perform each step of the inference chain?}

Table~\ref{tab:rq1_results} summarizes step-by-step performance across the inferential chain. \textbf{Overall, LLMs perform comparably to or better than human third-person observers on appraisal rating, emotion intensity prediction, and negative emotion label identification, while falling short on appraisal reasoning and positive emotion label identification.} For positive emotion label identification, models tend to exhibit greater imbalance between precision and recall compared to human observers, resulting in substantially lower Example F1 scores. For core appraisal reasoning, we combine automatic metrics with human evaluation for a comprehensive assessment: most models score below the human third-person observer on automatic metrics and best-response selection, yet achieve higher consistency scores in human evaluation — a divergence reflecting that human observers tend to draw on personal experience and produce more concise responses, making them more likely to be preferred when the reasoning is relatively unambiguous, whereas models tend toward thorough situational interpretation and generate richer responses, which are more frequently selected when partial alignment is present (see Appendix~\ref{app:rq1} for full human evaluation results). 

Across models, Claude and GPT consistently rank at the top across most tasks, including appraisal reasoning. Domain-specific fine-tuned models, however, present a notable dissociation: despite PsyLLM's strength on appraisal rating and EmoLLaMa's lead on appraisal reasoning metrics, neither shows a corresponding advantage on downstream emotion label selection, \textbf{suggesting that upstream inferential chain gains do not necessarily transfer to downstream tasks, and that strong downstream performance does not imply correct appraisal-mediated reasoning.}

\begin{table*}[ht]
\centering
\caption{Step-by-step evaluation results across the inferential chain, with first-person annotations as ground truth. Third-person human performance is averaged over all third-person annotations on the 500-sample subset. Results for open-source models are averaged over 5 runs; all standard deviations are $<$1\% and omitted. Consistency ratio (Cons.) and Best-response ratio (Best) are based on human evaluation on a 100-scenario sample. Best results among LLMs are in \textbf{bold}; second-best are \underline{underlined}. \colorbox{worse}{Red} cells indicate worse performance than Human. }
\label{tab:rq1_results}
\resizebox{\textwidth}{!}{%
\begin{tabular}{l ccc cc c cccc cccc}
\toprule
& \multicolumn{5}{c}{\textbf{Appraisal Reasoning}} 
& \multicolumn{1}{c}{\textbf{Appraisals}} 
& \multicolumn{4}{c}{\textbf{Positive Emotion}} 
& \multicolumn{4}{c}{\textbf{Negative Emotion}} \\
\cmidrule(lr){2-6} \cmidrule(lr){7-7} \cmidrule(lr){8-11} \cmidrule(lr){12-15}
Model
& BLEU & ROUGE-L & \makecell{BERT\\Score}
& \makecell{Cons.} & \makecell{Best}
& RMSE
& \makecell{Intensity\\RMSE} & \makecell{Example\\F1} & \makecell{Micro\\F1} & \makecell{Macro\\F1}
& \makecell{Intensity\\RMSE} & \makecell{Example\\F1} & \makecell{Micro\\F1} & \makecell{Macro\\F1} \\

\midrule
Third-Person    & 0.007 & 0.163 & 0.875 & 0.448 & 0.152 & 0.389 & 0.259 & 0.615 & 0.525 & 0.470 & 0.324 & 0.564 & 0.543 & 0.449 \\
\midrule
Claude   & 0.007 & \cellcolor{worse}\underline{0.157} & \cellcolor{worse}0.872 & \textbf{0.570} & \cellcolor{worse} \textbf{0.149} & \underline{0.357} & 0.213 & \cellcolor{worse}0.490 & \textbf{0.535} & \textbf{0.499} & \textbf{0.237} & \textbf{0.639} & \textbf{0.621} & \textbf{0.513} \\
GPT      & 0.008 &  0.156 & \cellcolor{worse}0.869 & \underline{0.569} & \cellcolor{worse} 0.121 & 0.365 & \textbf{0.210} & \cellcolor{worse}\textbf{0.556} & \underline{0.517} & \underline{0.475} & \underline{0.251} & \underline{0.630} & \underline{0.614} & \underline{0.500} \\
Qwen     & \cellcolor{worse}0.002 & \cellcolor{worse}0.144 & \underline{0.875} & 0.478 & \cellcolor{worse}0.083 & \cellcolor{worse}0.430 & 0.222 & \cellcolor{worse}\underline{0.513} & 0.509 & 0.466 & \underline{0.251} & 0.617 & 0.607 & \underline{0.511} \\
Llama    & \underline{0.009} & \cellcolor{worse}0.139 & \cellcolor{worse}0.866 & 0.504 & \cellcolor{worse}0.104 & \cellcolor{worse}0.395 & 0.246 & \cellcolor{worse}0.450 & \cellcolor{worse}0.463 & \cellcolor{worse}0.403 & \underline{0.251} & 0.601 & 0.578 & 0.458 \\
PsyLLM   & \cellcolor{worse}0.005 & \cellcolor{worse}0.138 & \cellcolor{worse}0.863 & 0.482 & \cellcolor{worse}0.094 & \textbf{0.349} & 0.239 & \cellcolor{worse}0.441 & \cellcolor{worse}0.451 & \cellcolor{worse}0.394 & 0.284 & 0.593 & 0.570 & 0.464 \\
EmoLlama & \textbf{0.013} & \textbf{0.175} & \textbf{0.878} & 0.504 & \cellcolor{worse} \underline{0.131} & \cellcolor{worse}0.443 & 0.254 & \cellcolor{worse}0.224 & \cellcolor{worse}0.383 & \cellcolor{worse}0.322 & 0.274 & \cellcolor{worse}0.532 & \cellcolor{worse}0.522 & \cellcolor{worse}0.345 \\
\bottomrule
\end{tabular}%
}
\end{table*}

We further examine fine-grained results at the appraisal dimension and emotion category levels (Appendix~\ref{app:rq1}). On both appraisal reasoning and appraisal rating, models face the greatest difficulty on the \textit{accountability}, \textit{control}, and \textit{certainty} dimensions, while performing relatively well on \textit{relevance}. For emotion labels, \textit{disgust} is the most consistently underperformed negative category, while \textit{hopeful}, \textit{grateful}, and \textit{inspiration} are systematically missed in positive emotion identification, with recall substantially lagging precision. Interestingly, this pattern of results is consistent with recent work of the developmental psychology of appraisal~\cite{doan2025emotion}, which suggests that \textit{relevance} and \textit{congruence} are among the earliest appraisal dimensions acquired in children's emotion understanding, enabling basic differentiation of positive and negative emotions, whereas \textit{control} and \textit{accountability} emerge later in development, involving more complex causal attribution and social reasoning, and are critical for fine-grained differentiation among discrete emotion categories. LLMs' performance exhibits exhibit the same hierarchy: models perform relatively well on coarse-grained, valence-level emotion identification, but show systematic gaps in categories that require higher-order appraisal reasoning to differentiate — relatively strong on valence-level identification but systematically weaker where higher-order appraisal reasoning is required — \textbf{suggesting that current models may not yet have fully acquired the cognitive capacities needed for fine-grained emotion differentiation.}

\textbf{RQ2: Does the appraisal reasoning process help improve appraisal prediction and emotion recognition?}

Table~\ref{tab:rq2_full} reports performance changes under different input conditions relative to the scenario-only baseline, with Micro-F1 and Macro-F1 for emotion recognition further illustrated in Figure~\ref{fig:appendix_emotion_f1}. \textbf{After injecting first-person appraisal reasoning text, appraisal prediction improved significantly across all models. At the emotion recognition level, however, different metrics tell divergent stories.} Reasoning text broadens emotion category coverage by enriching situational context, leading to widespread recall improvements, but may simultaneously introduce over-inference that prevents precision from improving correspondingly, leaving Example-F1 gains largely non-significant while Micro-F1 and Macro-F1 show notable gains across most models. In comparison, directly injecting first-person appraisal ratings more effectively improved emotion intensity prediction, but fell short of reasoning text in emotion category coverage; when both types of information are combined, negative emotion intensity prediction improves further, while gains on other metrics vary across models.

\begin{table*}[ht]
\centering
\small
\caption{Performance change ($\Delta$) under different input settings relative to the scenario-only baseline, assessed via paired $t$-tests. \colorbox{green}{Green} indicates significant improvement ($p<.05$); \colorbox{red}{Red} indicates significant degradation ($p<.05$).}
\label{tab:rq2_full}
\resizebox{0.75\textwidth}{!}{
\begin{tabular}{l ccccc c cccc}
\toprule
& \multicolumn{1}{c}{\textbf{Appraisal}}
& \multicolumn{2}{c}{\textbf{Pos. Emotion}}
& \multicolumn{2}{c}{\textbf{Neg. Emotion}}
&
& \multicolumn{2}{c}{\textbf{Pos. Emotion}}
& \multicolumn{2}{c}{\textbf{Neg. Emotion}} \\
\cmidrule(lr){2-2} \cmidrule(lr){3-4} \cmidrule(lr){5-6} \cmidrule(lr){8-9} \cmidrule(lr){10-11}
\textbf{Model}
& \makecell{RMSE}
& \makecell{Intensity\\RMSE} & \makecell{Example\\F1}
& \makecell{Intensity\\RMSE} & \makecell{Example\\F1}
&
& \makecell{Intensity\\RMSE} & \makecell{Example\\F1}
& \makecell{Intensity\\RMSE} & \makecell{Example\\F1} \\
\cmidrule(lr){2-6} \cmidrule(lr){8-11}
& \multicolumn{5}{c}{\textbf{+ 1st-person Reasoning}}
&
& \multicolumn{4}{c}{\textbf{+ 1st-person Appraisals}} \\
\midrule
Claude   & \colorbox{green!30}{$-$.019} & $-$.008 & \colorbox{green!30}{$-$.021} & $-$.011 & $-$.009 && \colorbox{green!57}{$-$.047} & $-$.011 & \colorbox{green!30}{$-$.024} & \colorbox{green!40}{$+$.029} \\
GPT      & \colorbox{green!30}{$-$.022} & $-$.003 & $-$.012 & $-$.015 & $-$.003 && \colorbox{green!40}{$-$.032} & $-$.019 & \colorbox{green!30}{$-$.021} & $-$.013 \\
Qwen     & \colorbox{green!50}{$-$.036} & $-$.004 & $+$.011 & \colorbox{green!20}{$-$.010} & $-$.010 && \colorbox{green!30}{$-$.022} & \colorbox{green!40}{$+$.029} & \colorbox{green!20}{$-$.012} & $+$.008 \\
Llama    & \colorbox{green!60}{$-$.052} & $-$.011 & $+$.008 & $-$.018 & $+$.005 && \colorbox{green!60}{$-$.047} & \colorbox{green!90}{$-$.079} & \colorbox{green!30}{$-$.023} & $+$.007 \\
PsyLLM   & \colorbox{green!10}{$-$.002} & \colorbox{red!10}{$+$.009} & $+$.018 & \colorbox{green!20}{$-$.012} & \colorbox{red!40}{$-$.029} && \colorbox{green!40}{$-$.030} & $-$.000 & \colorbox{green!50}{$-$.035} & $-$.006 \\
EmoLlama & \colorbox{green!20}{$-$.009} & $-$.002 & $+$.002 & $-$.005 & $+$.009 && $-$.013 & \colorbox{green!70}{$+$.059} & $+$.007 & \colorbox{green!30}{$+$.018} \\
\midrule
& \multicolumn{5}{c}{\textbf{+ Predicted Reasoning}}
&
& \multicolumn{4}{c}{\textbf{+ 1st-person Reasoning \& Appraisals}} \\
\midrule
Claude   & \colorbox{red!30}{$+$.017} & \colorbox{red!20}{$+$.009} & $-$.019 & \colorbox{red!60}{$+$.048} & \colorbox{red!70}{$-$.055} && \colorbox{green!60}{$-$.051} & \colorbox{red!40}{$-$.028} & \colorbox{green!40}{$-$.028} & \colorbox{green!40}{$+$.023} \\
GPT      & \colorbox{red!20}{$+$.011} & \colorbox{green!10}{$-$.001} & \colorbox{red!50}{$-$.038} & \colorbox{red!60}{$+$.047} & \colorbox{red!100}{$-$.103} && \colorbox{green!0.40}{$-$.034} & $-$.022 & \colorbox{green!40}{$-$.030} & $-$.010 \\
Qwen     & \colorbox{red!30}{$+$.020} & \colorbox{red!40}{$+$.030} & $-$.007 & \colorbox{red!50}{$+$.044} & \colorbox{red!60}{$-$.048} && \colorbox{green!50}{$-$.037} & \colorbox{green!40}{$+$.028} & \colorbox{green!30}{$-$.024} & $+$.004 \\
Llama    & \colorbox{green!40}{$-$.033} & \colorbox{red!30}{$+$.021} & \colorbox{green!40}{$+$.032} & \colorbox{red!30}{$+$.025} & \colorbox{red!80}{$-$.073} && \colorbox{green!80}{$-$.067} & \colorbox{red!80}{$-$.068} & \colorbox{green!40}{$-$.030} & $+$.015 \\
PsyLLM   & \colorbox{red!30}{$+$.022} & \colorbox{red!50}{$+$.035} & $+$.014 & \colorbox{green!30}{$-$.021} & \colorbox{red!50}{$-$.039} && $-$.023 & $+$.006 & \colorbox{green!50}{$-$.040} & \colorbox{red!50}{$-$.040} \\
EmoLlama & $+$.001 & \colorbox{green!30}{$-$.019} & $+$.014 & \colorbox{red!10}{$+$.001} & $-$.017 && $-$.016 & \colorbox{green!40}{$+$.027} & $+$.002 & $+$.009 \\
\bottomrule
\end{tabular} 
}
\end{table*}

In contrast, injecting model-predicted appraisal reasoning yielded a markedly different pattern: appraisal and emotion intensity prediction degraded significantly across most models, and negative emotion Example-F1 dropped substantially, with several models showing a decline in negative emotion Micro-F1 even relative to baseline. Qualitative analysis further reveals that models tend to employ exaggerated emotional language in their generated reasoning texts, which may represent one mechanism by which self-generated reasoning introduces noise that propagates to downstream emotion inference. \textbf{Taken together, while LLMs can leverage first-person appraisal reasoning to improve performance, their own capacity to generate reasoning remains limited.}

Dimension-specific appraisal reasoning injection results are provided in Appendix~\ref{app:rq2}. For appraisal prediction, injecting \textit{accountability}, \textit{control}, and \textit{certainty} reasoning consistently improved predictions on corresponding sub-dimensions across most models, while the effects of \textit{congruence} and \textit{relevance} were comparatively weak, suggesting that supplementing reasoning on dimensions where models are weak yields the substantial gains. The same tendency holds for emotion recognition, where reasoning effects are likewise concentrated on emotion categories where models underperform at baseline. \textbf{Thus, accurate appraisal reasoning information can selectively compensate for models' deficiencies in specific appraisal dimensions and emotion inference.}

\textbf{RQ3: Are emotion predictions in LLMs sensitive to structured appraisal interventions?}

\begin{figure}[ht]
\centering
\includegraphics[width=\columnwidth]{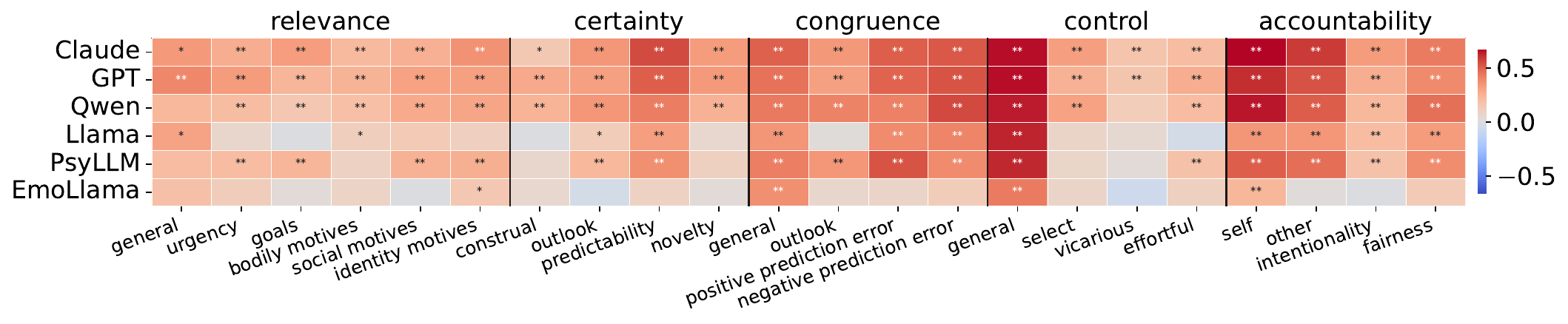}
\caption{Pearson correlation between model-predicted rating changes and human annotation changes under causal intervention. Red indicates positive correlation; $*p<.05$, $**p<.01$.}
\label{fig:rq3_exp2}
\end{figure}

Figure~\ref{fig:rq3_exp2} presents the intervention results for $r\rightarrow\mathbf{a}$; full heatmaps for $r\rightarrow e_i$ and $r\rightarrow e_l$ are provided in Appendix~\ref{app:rq3}. For $r\rightarrow\mathbf{a}$, Claude, GPT, and Qwen show moderate-to-strong positive correlations across most appraisal dimensions, while PsyLLM and Llama perform more weakly, and EmoLLaMa shows near-zero response. For $r\rightarrow e_i$, most models exhibit strong and consistent sensitivity, with the exception of EmoLLaMa; correlations are strongest for \textit{congruence}, consistent with theoretical predictions. For $r\rightarrow e_l$, the strongest correlations across all models concentrate on \textit{relevance.general} and \textit{congruence}, aligning with appraisal theory's core predictions regarding the role of event relevance and goal congruence in determining emotion valence. Claude, GPT, and Qwen show moderate-to-strong correlations for negative emotion labels across most dimensions, whereas positive emotion responses are consistently weaker across all models, corroborating RQ1's finding that positive emotion prediction is more challenging. Llama and PsyLLM achieve reasonable negative emotion prediction in RQ1, yet show notably weaker directional sensitivity than Claude, GPT, and Qwen, indicating that downstream performance does not reliably reflect sensitivity to appraisal structure. \textbf{Overall, emotion prediction performance alone does not imply acquired appraisal reasoning, but most models exhibit theory-consistent sensitivity patterns on key dimensions, with stronger models showing broader and more pronounced sensitivity.}

\textbf{RQ4: Do LLMs capture the distributional heterogeneity of human appraisal judgments?}

\begin{table*}[ht]
\centering
\small
\caption{Subjectivity modeling results. Mean MAE, Variance MAE, and Wasserstein distance measure rating distribution fitting quality (lower is better); Spearman $\rho$ and Precision@$k$ measure dimension disagreement pattern reproduction (higher is better).}
\label{tab:rq4}
\resizebox{\textwidth}{!}{%
\begin{tabular}{l l ccc | cccc c}
\toprule
& & \multicolumn{3}{c|}{\textbf{Rating Distribution Fitting}} 
& \multicolumn{5}{c}{\textbf{Dimension Disagreement Pattern}} \\
\cmidrule(lr){3-5} \cmidrule(lr){6-10}
& & Mean MAE $\downarrow$ & Variance MAE $\downarrow$ & Wasserstein $\downarrow$ 
& Spearman $\rho$ $\uparrow$ & P@1 $\uparrow$ & P@3 $\uparrow$ & P@5 $\uparrow$ & P@10 $\uparrow$ \\
\midrule
\multirow{4}{*}{Qwen}
& Repeated sampling   & 0.936 & 0.816 & 0.992 & $+$0.204 & 0.054 & 0.157 & 0.275 & 0.536 \\
& $+$ Persona         & 0.874 & \textbf{0.772} & 0.935 & $+$0.223 & 0.078 & 0.167 & 0.278 & 0.544 \\
& $+$ Core appraisals & 0.758 & 0.875 & 0.852 & $+$0.271 & 0.068 & 0.185 & \textbf{0.298} & \textbf{0.565} \\
& $+$ Both            & \textbf{0.729} & 0.806 & \textbf{0.813} & $+$\textbf{0.282} & \textbf{0.086} & \textbf{0.195} & 0.295 & 0.556 \\
\midrule
\multirow{4}{*}{Llama}
& Repeated sampling   & 0.819 & 0.858 & 0.978 & $+$0.087 & 0.046 & 0.166 & 0.264 & 0.483 \\
& $+$ Persona         & 0.895 & 0.890 & 1.087 & $-$0.061 & 0.036 & 0.139 & 0.223 & 0.435 \\
& $+$ Core appraisals & \textbf{0.666} & \textbf{0.786} & \textbf{0.837} & $+$\textbf{0.127} & \textbf{0.064} & \textbf{0.182} & \textbf{0.281} & \textbf{0.511} \\
& $+$ Both            & 0.785 & 0.841 & 0.981 & $+$0.018 & 0.036 & 0.161 & 0.260 & 0.467 \\
\midrule
\multirow{4}{*}{PsyLLM}
& Repeated sampling   & 0.697 & 0.833 & 0.908 & $+$0.080 & 0.054 & 0.156 & 0.244 & 0.481 \\
& $+$ Persona         & 0.750 & 0.828 & 0.960 & $+$0.031 & 0.028 & 0.126 & 0.219 & 0.471 \\
& $+$ Core appraisals & \textbf{0.633} & \textbf{0.734} & \textbf{0.763} & $+$\textbf{0.230} & 0.068 & \textbf{0.189} & \textbf{0.290} & \textbf{0.539} \\
& $+$ Both            & 0.665 & 0.751 & 0.818 & $+$0.136 & \textbf{0.086} & 0.184 & 0.268 & 0.498 \\
\midrule
\multirow{4}{*}{EmoLlama}
& Repeated sampling   & 1.102 & 0.892 & 1.245 & $-$\textbf{0.063} & 0.034 & 0.119 & \textbf{0.201} & \textbf{0.443} \\
& $+$ Persona         & 1.098 & \textbf{0.886} & 1.246 & $-$0.089 & 0.034 & 0.107 & 0.184 & 0.426 \\
& $+$ Core appraisals & \textbf{1.046} & 0.898 & \textbf{1.227} & $-$0.119 & \textbf{0.038} & \textbf{0.122} & 0.199 & 0.425 \\
& $+$ Both            & 1.060 & 0.906 & 1.255 & $-$0.118 & 0.030 & 0.116 & 0.195 & 0.432 \\
\bottomrule
\end{tabular}%
}
\end{table*}

Table~\ref{tab:rq4} reports results on models' ability to approximate human appraisal rating distributions and reproduce inter-annotator disagreement patterns. For distribution fitting, injecting core appraisal reasoning yields the most consistent improvements across all models, with notable reductions in mean MAE and Wasserstein distance; persona injection shows unstable effects, and combining both types is not always optimal. For disagreement pattern reproduction, Spearman $\rho$ and $P@k$ remain close to random baseline, with consistent but modest improvements from reasoning injection and little benefit from persona. Further analysis of output variance reveals that reasoning injection substantially increases variance in most models, yet $\sigma$-MAE and disagreement reproduction remain well below human levels, suggesting that models can produce differentiated ratings but through mechanisms that likely differ from human inter-annotator disagreement. \textbf{In sum, while external appraisal reasoning provides meaningful support for modeling subjective heterogeneity, current models have not internalized the mechanisms necessary to capture inter-observer differences autonomously.}

\section{Conclusion and Future Work}

This paper presents a process-level evaluation framework for LLM emotion understanding grounded in appraisal theory, and introduces the first dataset with complete dual-perspective inferential chain annotations covering both first- and third-person perspectives on real-world narratives. Through systematic experiments across six LLMs, we find that LLMs achieve or surpass human observer performance on certain steps of the inferential chain, yet exhibit consistent shortfalls in appraisal reasoning generation and positive emotion recognition. Furthermore, while first-person reasoning has a substantial effect on model understanding, model-generated reasoning fails to improve and not all models show sensitivity to appraisal reasoning intervention. Additionally, existing models remain unable to capture the distributional variation in emotion understanding observed across human annotators. These findings suggest that evaluating LLMs along the full inferential chain reveals capability gaps that emotion prediction metrics alone would miss, offering a more diagnostically informative foundation for assessing LLMs' emotion understanding.

Future work may proceed along several directions. First, the present framework characterizes model capabilities along the inferential chain at the behavioral level, while future work may apply mechanistic interpretability methods to investigate how appraisal reasoning is encoded in models' internal representations. Second, the multi-perspective and multi-label annotation structure of the dataset provides a foundation for modeling emotion as a situation-conditioned distribution, opening up avenues for learning continuous emotion embeddings that better capture subjective heterogeneity in emotion understanding. Third, several limitations identified in this work, including insufficient quality of model-generated reasoning, consistently weak positive emotion recognition, and the failure to capture distributional variation across human annotators, warrant attention in future efforts to improve LLM emotion understanding through training and alignment. Finally, future work may also explore transferring the dataset and framework to downstream emotion support tasks, such as leveraging the complete inferential chain annotations to support the development and evaluation of cognitive reappraisal-based emotion regulation systems.

\section{Limitations}
\label{sec:limitations}
While the data collection pipeline was designed and reviewed in close collaboration with experienced psychology researchers to ensure the validity of the annotation schema, several limitations remain. First, data are drawn exclusively from English-speaking participants in the United States and the United Kingdom, meaning that appraisal patterns and emotion vocabulary are inevitably shaped by Anglo-American cultural contexts, and the generalizability of our findings to other languages and cultures remains to be validated. Furthermore, while the micro-narratives framework reduces the cognitive burden on participants, it introduces the risk of LLM-generated reconstructions incorporating extraneous details, though we mitigated this through extensive prompt refinement. In addition, participants' self-selection of narrative events leads to a pronounced imbalance between negative and positive emotion samples, which somewhat limits the coverage of the dataset. Finally, due to the high cost of annotation, the third-person annotation set comprises only 500 narratives with five annotators each, which constrains the precision of distributional estimates for subjective variation.

\begin{ack}


This work was supported in part by the UK Engineering and Physical Sciences Research Council through a Turing AI Fellowship (grant no. EP/V020579/1, EP/V020579/2), a UKRI Future Leaders fellowship (grant no. MR/Y034295/1 and MR/T041897/1), the Prosperity Partnership scheme (grant no. UKRI566), the Estonian Research Council (grant PRG3182), and Estonian Ministry of Research and Education (grant TK218 Estonian Center of Excellence for Well-Being Sciences).

\end{ack}

\bibliography{references}
\bibliographystyle{unsrt}

\appendix
\setcounter{table}{0}
\renewcommand{\thetable}{A\arabic{table}}
\setcounter{figure}{0}
\renewcommand{\thefigure}{A\arabic{figure}}

\section{Technical Appendices}

\subsection{Related Work}

\paragraph{Appraisal Theory and Emotion Modeling}
Affective computing has long been dominated by discrete emotion categories (e.g., Ekman's six basic emotions~\cite{ekman1992argument}) and valence-arousal dimensional spaces~\cite{russell1980circumplex}, which fall short of capturing the richness of real-world emotional experience~\cite{zall2025intelligent}. More recently, appraisal theory has gained traction as a more structured alternative: emotions arise not directly from events, but from individuals' evaluations of situations along multiple dimensions such as goal relevance, controllability, and accountability~\cite{smith1985patterns, roseman1991appraisal, krajuvskina2025unpacking}. Appraisal process models further characterize appraisal as a multi-stage sequential cognitive process~\cite{scherer2009dynamic, marsella2009ema}. Inspired by appraisal process models, our evaluation framework explicitly incorporates the reasoning step into the inferential chain and integrates multiple theoretical accounts of appraisal and emotion into a unified structure, enabling systematic examination of whether LLMs' emotion-related behaviors align with the predictions of established psychological theories.

\begin{table}[h]
\centering
\resizebox{\textwidth}{!}{%
\begin{tabular}{lcccccccccc}
\toprule
\textbf{Dataset} & \makecell{\textbf{Data}\\\textbf{Source}} & \makecell{\textbf{Appraisal}\\\textbf{Reasoning}} & \makecell{\textbf{Appraisal}\\\textbf{Rating}} & \makecell{\textbf{Emotion}\\\textbf{Label}} & \makecell{\textbf{Emotion}\\\textbf{Intensity}} & \textbf{Perspective} & \makecell{\textbf{Multiple}\\\textbf{Annotators}} & \makecell{\textbf{\#}\\\textbf{Instances}} & \makecell{\textbf{Avg. Text}\\\textbf{Length}} \\
\midrule
Crowd-Envent~\cite{troiano2023dimensional}     & human     & \xmark & \cmark & single   & \cmark & dual & \cmark & 6,600 & 20 \\
CovidET-Appraisals~\cite{zhan2023evaluating}   & human     & \cmark & \cmark & multiple & \xmark & 3rd  & \xmark & 241 & 175 \\
PEACE-Reviews~\cite{yeo2023peace}              & human     & \xmark & \cmark & multiple & \cmark & 3rd  & \xmark & 1,400 & 150  \\
CAPE~\cite{liu2025cape}                        & synthetic & \cmark & \xmark & single   & \xmark & 3rd  & \xmark & 2,848 & 385\\
Yeo et al. (2025)~\cite{yeo2025beyond}         & synthetic & \cmark & \cmark & ---      & \xmark & 3rd  & \xmark & 432 & $\sim$150\\
CoRE~\cite{bhattacharyya2026do}                & synthetic & \cmark & \cmark & single   & \xmark & 1st  & \xmark & 308 & - \\
\textbf{CAREBench}                             & human     & \cmark & \cmark & multiple & \cmark & dual & \cmark & 1,000 & 168 \\
\bottomrule
\end{tabular}%
}
\caption{Comparison of appraisal-annotated datasets. }
\label{tab:dataset_comparison}
\end{table}

\paragraph{Appraisal-Annotated Datasets}
Several datasets have been constructed to ground emotion annotation in appraisal theory. Table~\ref{tab:dataset_comparison} summarizes their key characteristics in comparison to ours. Specifically, Troiano et al.~\cite{troiano2023dimensional} built a dual-perspective appraisal corpus (Crowd-Envent) comprising both author self-reports and reader reconstructions, demonstrating that appraisal dimensions can be reliably predicted from text and improve emotion classification. However, it consists solely of appraisal rating scales without capturing the underlying reasoning process. CovidET-Appraisals~\cite{zhan2023evaluating} annotated 24 appraisal dimensions with natural language rationales on Reddit posts about Covid-19, though the dataset is limited in scale (241 posts) and covers only a single perspective with single-annotator labels. PEACE-Reviews~\cite{yeo2023peace} provides appraisal ratings and multi-label emotion annotations over post-purchase reviews, yet similarly lacks appraisal reasoning and covers only the third-person perspective. Another line of work introduces appraisal annotations over synthetically generated scenarios, including LLM-generated Chinese emotional dialogues~\cite{liu2025cape}, model-generated scenarios with appraisal reasoning~\cite{bhattacharyya2026do}, and a forward and backward appraisal-emotion reasoning benchmark based on Prisoner's Dilemma scenarios~\cite{yeo2025beyond}; as these datasets are not grounded in authentic human experience, their annotations can only be validated against theoretical predictions rather than reflecting genuine individual appraisal patterns. In contrast, our dataset is grounded in real emotional narratives and, to our knowledge, is the first to provide complete inferential chain annotations from both first-person and third-person perspectives — spanning open-ended appraisal reasoning texts, hierarchical appraisal ratings, and multi-label emotion annotations with polarity — enabling process-level evaluation of LLMs' emotion understanding.

\paragraph{Appraisal-Grounded LLM Emotion Understanding Evaluation} 
Several existing studies evaluating LLMs' appraisal capabilities have built on the Crowd-Envent dataset, finding that GPT-4 performs as a reasonable appraisal annotator comparable to human readers~\cite{ruder2025assessing}, but tends to align more closely with third-person observer judgments than first-person self-attributions~\cite{tak2024gpt}; Debnath et al.~\cite{debnath2025appraisal} further incorporate appraisal as an intermediate representation in dialogue emotion modeling, demonstrating its cross-corpus transferability, while Tak et al. reveal through probing analyses that hierarchical appraisal signals are present within LLMs' internal representations~\cite{tak2025mechanistic}. Gandhi et al.~\cite{gandhi2024human} extend this line of work to a broader set of models (GPT-4, Claude-3, Gemini-1.5-Pro) in controlled settings, also finding that model judgments can surpass human performance in stereotypical scenarios, though their stimuli are synthetically generated and cover only single emotion labels, limiting ecological validity. Yeo \& Jaidka~\cite{yeo2025beyond} and Yongsatianchot et al.~\cite{yongsatianchot2023investigating} evaluate LLMs' appraisal reasoning in constrained settings — a Prisoner's Dilemma paradigm and a psychological stress questionnaire, respectively — using theoretical predictions as the comparison baseline. However, none of the above work accounts for individual differences in emotional understanding. Zhou et al.~\cite{zhou2025modeling} address this gap by modeling subjectivity in appraisal judgments as distributions, finding that LLMs fall short in capturing inter-individual appraisal variance. Building on these efforts, our work provides a systematic evaluation spanning the full inferential chain from appraisal reasoning to multi-emotion recognition, while jointly covering first-person emotion understanding and third-person distributional heterogeneity modeling.

\subsection{Data Collection Details}
\label{app:data-collection}

\paragraph{Scenario Collection}
Scenarios were collected using the micro-narratives framework, in which participants respond to a series of sub-questions guided by an LLM, whose responses are then reorganized and minimally reformulated by the LLM into a coherent narrative. The framework aims to reduce the linguistic burden of generating extended text while preserving participants' own words and voice as closely as possible \cite{skeggs2025micro}. The sub-questions for scenario collection were: (1) What happened? (2) What were you trying to do? What did you want or need in this situation? (3) How did the situation leave you feeling? (4) Is there anything else that was important to you in this situation? The questions were designed to capture the aspects of the situation most salient to the participant and most likely to have elicited their emotional response, and were finalized through review of pilot study data and discussion with psychology researchers.

\paragraph{Appraisal Inference and Ratings}
 
For each narrative, we collected open-ended inference texts for five core appraisal dimensions — relevance, certainty, congruence, control, and accountability — and subsequently Likert ratings for 22 appraisal items derived through hierarchical expansion of these dimensions. This design draws on Krajuškina et al.\cite{krajuvskina2025unpacking}, who note that the five core dimensions are included in the vast majority of appraisal theories, capture the broad aspects of situational appraisal, and yield the most stable predictions of affective valence; granular sub-dimensions, in turn, serve as specific instantiations of the broad dimensions, reflecting the particular reasons why a situation may be appraised in a given way along a broad dimension, and can capture affective variation that broad dimension scores alone cannot explain. Accordingly, we use inference texts to cover the core dimensions and complement them with rating scores at the granular level. 

Core appraisal inference texts were likewise collected using the micro-narratives framework, with one guided question per core dimension. Each question was designed to elicit the annotator's interpretation of the situation along the corresponding dimension. The definitions of the five dimensions and their associated questions are presented below.

\begin{tcolorbox}[
    enhanced,
    breakable,
    colback=gray!5!white,
    colframe=gray!30!white,
    boxrule=0.5pt,
    left=8pt, right=8pt, top=6pt, bottom=6pt,
    fontupper=\small
]
\begin{description}[leftmargin=0pt, itemsep=4pt]
    \item[\textbf{Relevance}]: the importance of the situation for the annotator's salient goals, needs, and motives.\\
    \textit{``When the event happened, how much did it matter to you? Why?''}

    \item[\textbf{Certainty}]: the clarity of the situation and the predictability of its implications.\\
    \textit{``Did things go as you expected? If not, what was unexpected?''}

    \item[\textbf{Congruence}]: the extent to which the situation facilitates or obstructs the pursuit of one's motives.\\
    \textit{``In what ways did this situation help or hurt you?''}

    \item[\textbf{Control}]: the perceived ability to change the situation or adapt to its effects.\\
    \textit{``Did you feel like you were in control? Why?''}

    \item[\textbf{Accountability}]: the perceived locus of responsibility for the situation.\\
    \textit{``Who or what did you feel was most responsible for all of this?''}
\end{description}
\end{tcolorbox}

Appraisal ratings were collected using 22 items derived through hierarchical expansion of the five core dimensions. Each item is presented as a statement, which annotators rated on a 5-point Likert scale based on their perception of the situation. The item indices, corresponding dimensions, and statements are listed in Table~\ref{tab:appraisal-items}.

\begin{table}[h]
\centering
\small
\caption{Appraisal rating statements grouped by core dimension.}
\begin{tabular}{llp{7cm}}
\toprule
\textbf{ID} & \textbf{Dimension} & \textbf{Statement} \\
\midrule
\multicolumn{3}{l}{\textit{Relevance}} \\
A1 & General & This situation matters to me. \\
A2 & Urgency & I need to do something about this situation right away. \\
A3 & Goals & This situation relates to my goals and plans. \\
A4 & Bodily motives & This situation relates to my physical well-being. \\
A5 & Social motives & This situation involves people who matter to me. \\
A6 & Identity motives & This situation concerns who I am and what I stand for. \\
\midrule
\multicolumn{3}{l}{\textit{Certainty}} \\
A7 & Construal & It is clear to me what is going on in this situation. \\
A8 & Outlook & I know what will come next in this situation. \\
A9 & Predictability & I saw this situation coming. \\
A10 & Novelty & This is a new kind of situation for me. \\
\midrule
\multicolumn{3}{l}{\textit{Congruence}} \\
A11 & General & This is a good situation. \\
A12 & Outlook & This situation will get better with time. \\
A13 & Positive prediction error & This situation is better than I expected. \\
A14 & Negative prediction error & This situation is worse than I expected. \\
\midrule
\multicolumn{3}{l}{\textit{Control}} \\
A15 & General & This situation is under my control. \\
A16 & Selectability & I can decide whether to stay in this situation or leave it. \\
A17 & Vicarious & Someone can handle this situation for me. \\
A18 & Effortful & I have to exert effort in this situation. \\
\midrule
\multicolumn{3}{l}{\textit{Accountability}} \\
A19 & Self & I am responsible for this situation. \\
A20 & Other & Someone else is responsible for this situation. \\
A21 & Intentionality & This situation was caused intentionally. \\
A22 & Fairness & This situation is fair and deserved. \\
\bottomrule
\end{tabular}
\label{tab:appraisal-items}
\end{table}

\paragraph{Emotion Intensity and Labels}
Emotion intensity was rated separately for positive and negative affect on a 7-point Likert scale, ranging from ``Not at all'' to ``Extremely.'' Emotion labels were selected from two predefined sets of ten categories each, covering positive and negative emotions respectively, with multiple selections and co-occurring polarities permitted. Each category is represented by a group of closely related terms to accommodate individual variation in emotion verbalization; in the main text, we refer to each category by its first term for brevity.

\begin{tcolorbox}[
  colback=gray!8,
  colframe=gray!40,
  boxrule=0.4pt,
  arc=2pt,
  left=6pt, right=6pt, top=4pt, bottom=4pt,
  fontupper=\small,
  breakable
]
\begin{tabularx}{\linewidth}{XX}
\textcolor{blue}{\textbf{Positive Labels}} & \textcolor{red}{\textbf{Negative Labels}} \\[3pt]
Hopeful, optimistic, encouraged     & Angry, frustrated, annoyed \\
Grateful, appreciative, thankful    & Worried, nervous, fearful \\
Glad, happy, joyful                 & Sad, downhearted, unhappy \\
Love, closeness, trust              & Disgust, distaste, revulsion \\
Amused, entertained                 & Despair, hopelessness, sorrow \\
Sympathy, concern, compassion       & Ashamed, humiliated, embarrassed \\
Inspiration, admiration, appreciation & Lonely, isolated, disconnected from others \\
Calm, peaceful, relaxed             & Panicked, alarmed, freaked out \\
Proud, confident, determined        & Guilty, blameworthy, repentant \\
Excited, enthusiastic, elated       & Confused, disoriented, surprised \\
\end{tabularx}
\end{tcolorbox}

\paragraph{Quality Control}

To ensure data quality, we conducted multiple rounds of pilot studies prior to the main data collection, iteratively refining the procedure and instructions based on participant feedback in collaboration with psychology experts. For formal data collection, participants were required to reside in the United Kingdom or the United States, with English as both their first and primary language, fluent English proficiency, a Prolific approval rate above 90\%, and more than 50 prior submissions on the platform. To guard against AI-generated responses, we implemented paste detection, typing speed anomaly detection, and response length anomaly detection, with flagged submissions subject to manual review and removal if suspected to be AI-generated. In addition, attention check items were embedded in the appraisal questionnaire, and submissions that failed these checks were excluded. Following data collection, we further removed submissions flagged as suspected bots by the Qualtrics platform, as well as those with unusually short completion times (under 5 minutes for first-person annotations or under 15 minutes for third-person annotations) or excessively long conversation turns (more than 3 turns beyond the designated questions).

\subsection{Annotator Demographics}
\label{app:annotator-demo}

This section reports the demographic characteristics of annotators in our dataset. During recruitment, we applied gender-balanced sampling to ensure comparable gender composition across the first-person and third-person groups.

\begin{figure}[ht]
    \centering

    \begin{subfigure}[b]{\textwidth}
        \centering
        \includegraphics[width=0.8\textwidth]{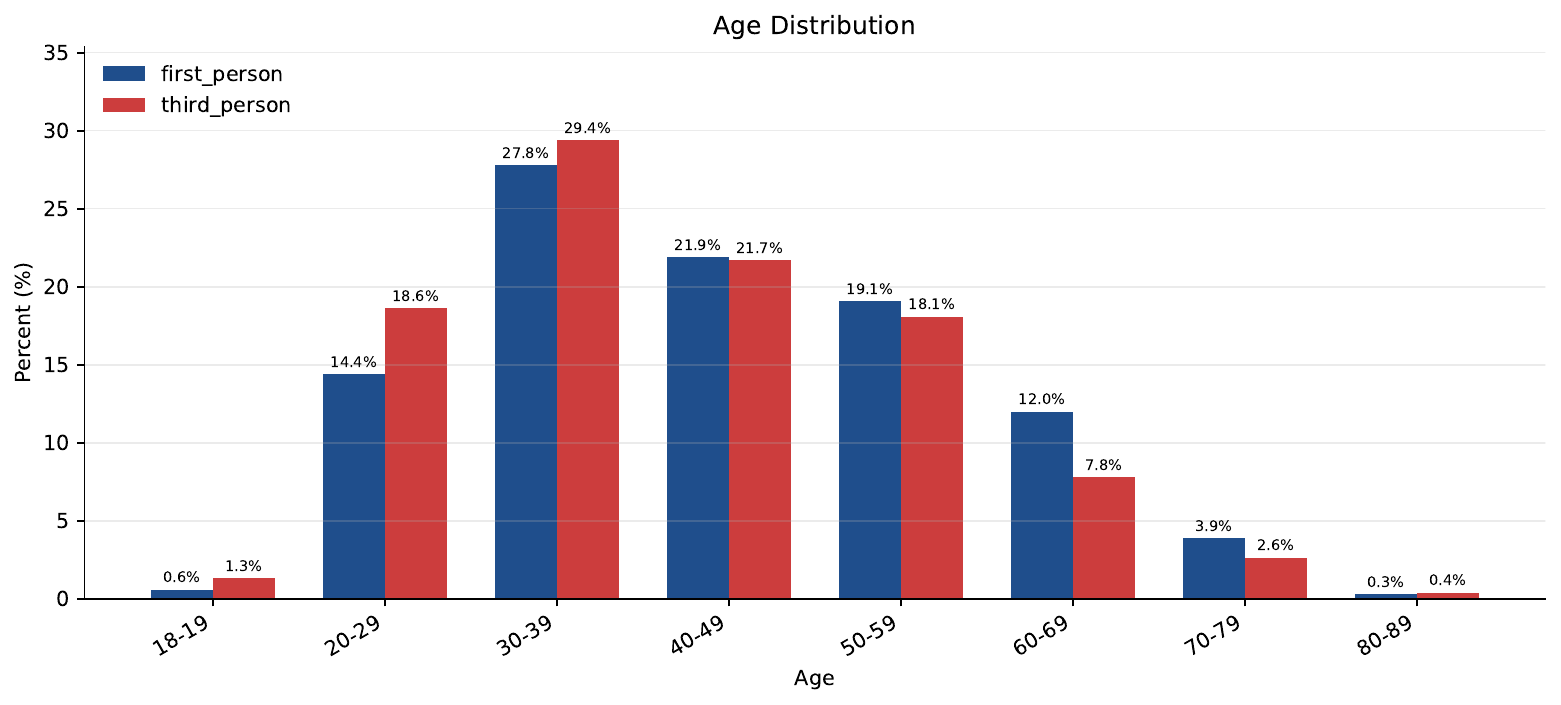}
        \caption{Age distribution.}
        \label{fig:age_distribution}
    \end{subfigure}
    
    \begin{subfigure}[b]{0.45\textwidth}
        \centering
        \includegraphics[width=\textwidth]{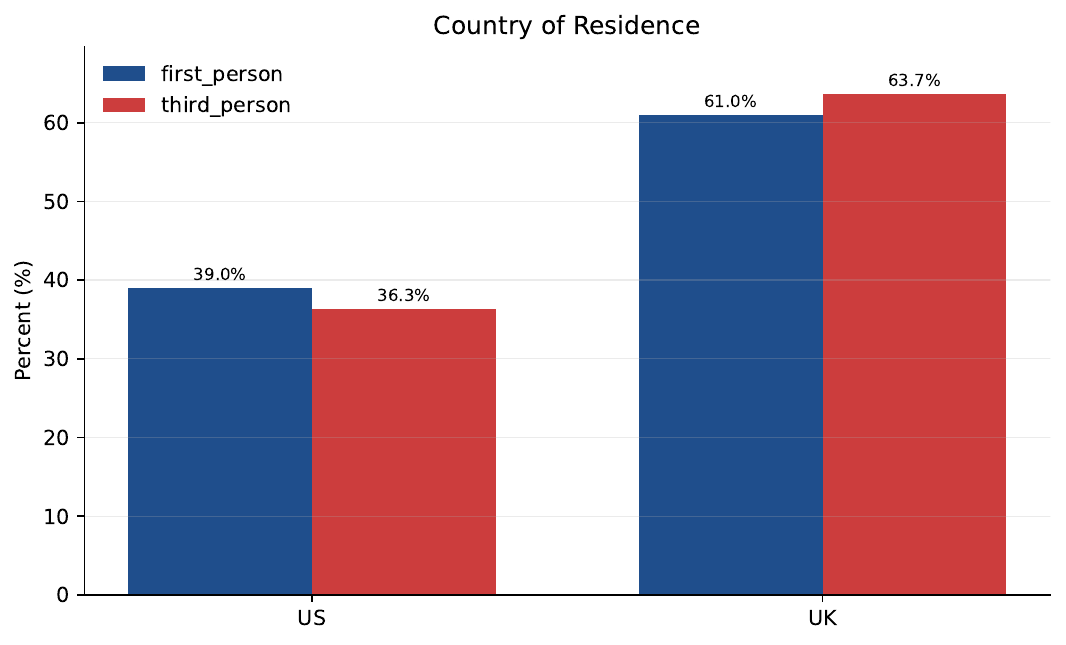}
        \caption{Country of residence distribution.}
        \label{fig:country_residence}
    \end{subfigure}
    \hfill
    \begin{subfigure}[b]{0.48\textwidth}
        \centering
        \includegraphics[width=\textwidth]{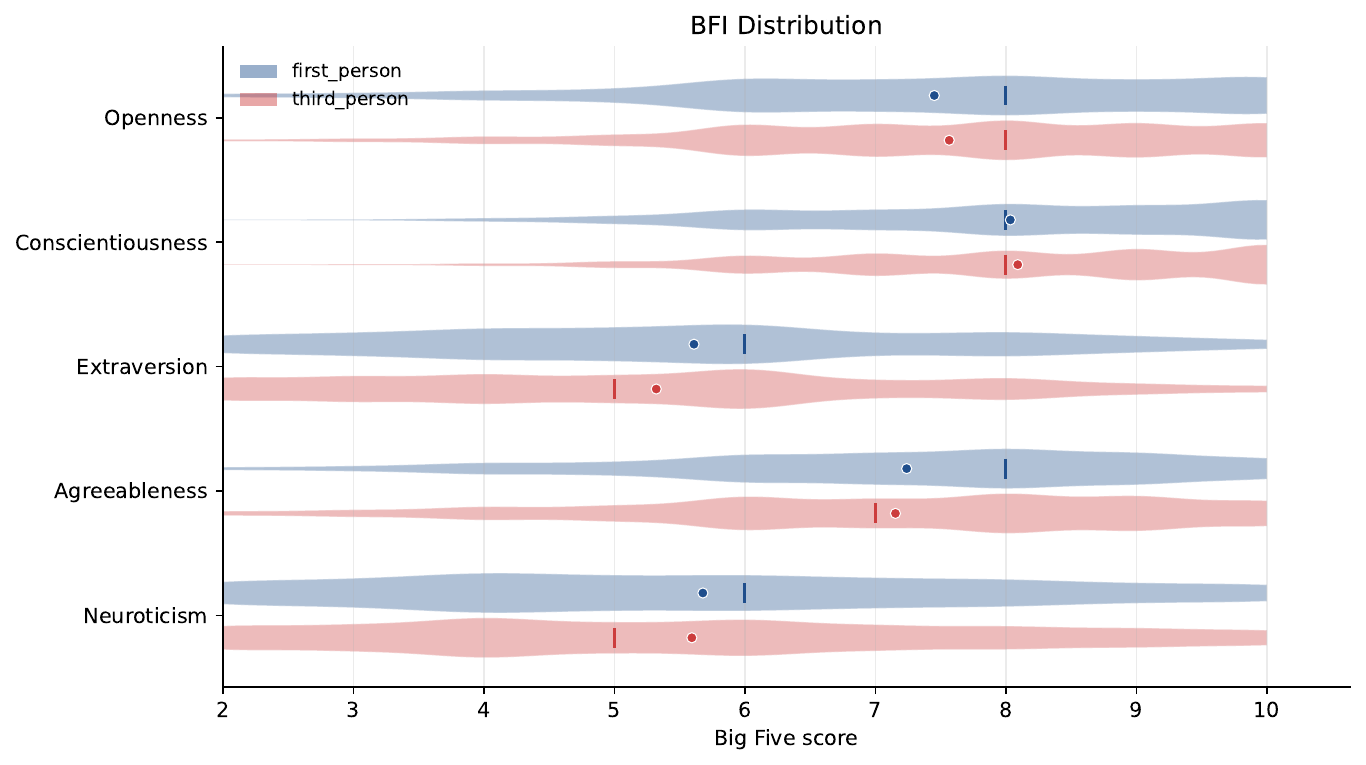}
        \caption{Big Five personality score distributions.}
        \label{fig:bfi_distribution}
    \end{subfigure}
    
    \vspace{1em}
    
    \caption{Demographic and personality characteristics of annotators in the first-person and third-person groups.}
    \label{fig:annotator_demographics}
\end{figure}

\paragraph{Age distribution.} As shown in Figure~\ref{fig:age_distribution}, the age distributions of the two groups are broadly similar, with both centered on the 30--39 age bracket, followed by the 20--29 and 40--49 groups. The samples span a wide age range, reflecting reasonable demographic diversity.

\paragraph{Country of residence.} To ensure comparable levels of English language proficiency, recruitment was restricted to residents of the UK and the US. As shown in Figure~\ref{fig:country_residence}, both groups are predominantly UK-based, and the geographic composition is closely matched across the two groups.

\paragraph{Big Five personality distribution.} As shown in Figure~\ref{fig:bfi_distribution}, the two groups exhibit broadly similar distributions across all five personality dimensions. Both groups score higher on Openness and Conscientiousness, while Extraversion and Neuroticism show relatively lower medians, and Agreeableness falls in between. The third-person group scores slightly lower on Extraversion and Neuroticism compared to the first-person group, though the overall differences are modest, indicating good comparability in personality composition between the two groups.

\subsection{Reproducibility Details}
\label{app:reprod}
\subsubsection{Supplementary Experimental Settings}
All experiments use a zero-shot setting, with each step evaluated independently. Every task prompt follows a uniform template comprising an instruction, the situation, a task description, and output constraints. Models are instructed to adopt the perspective of the story's narrator, read the first-person narrative carefully, and complete the corresponding task. The query format for each task mirrors that used by human annotators: for core-appraisal reasoning, models respond in open-ended text to the question associated with each appraisal dimension; for appraisal rating, models select their degree of agreement with the statement corresponding to each dimension on a 5-point Likert scale; for emotion intensity, models select from a 7-point scale the intensity of positive and negative emotion experienced in the described situation; and for emotion label, models select all applicable labels from a predefined set. The full prompt templates for each task are provided in Appendix~\ref{app:task-prompt}.

GPT-5.2 and Claude-Sonnet-4.6 are accessed via API at a total cost of approximately \$700; all open-source models are deployed on a local NVIDIA H100 80GB server and served via vLLM, with 10 concurrent threads, each model completing a single task run in approximately 30 minutes. For models supporting a thinking mode, this feature is disabled as our pilot study found no significant performance gain while inference cost increased substantially. For RQ4, temperature is set to 0.7 to encourage response diversity; all other experiments use a temperature of 0.2 to promote behavioral consistency.

\subsubsection{Task Prompts}
\label{app:task-prompt}

The task-specific prompts are shown below. The variable \texttt{\{scenario\}} denotes the situational narrative provided to the model, which depending on the experimental condition may be the story alone or the story augmented with core appraisal reasoning. The variable \texttt{\{question\}} denotes the dimension-specific question for each of the five core appraisal dimensions. The variable \texttt{\{statement\}} denotes the dimension-specific rating statement for each of the 22 appraisal items. For emotion labels, \texttt{\{emotion\_labels\}} refers to the predefined label sets described above.

\begin{tcolorbox}[
  colback=gray!8,
  colframe=gray!40,
  boxrule=0.4pt,
  arc=2pt,
  left=6pt, right=6pt, top=4pt, bottom=4pt,
  fontupper=\small\ttfamily,
  breakable,
  title={\small\bfseries Appraisal Reasoning}
]
\textbf{Instruction:} Imagine you are the person who wrote the following story. Read it carefully and internalize the feelings and situation described. Please answer the following questions in the first person. When answering, please think back to the moment when the events described in the story had just ended. Keep your response concise and try to keep it within 60 tokens.\\[3pt]
\textbf{My Situation:} \{scenario\}\\[3pt]
\textbf{Question:} \{question\}\\[3pt]
\textbf{My Answer:}
\end{tcolorbox}

\begin{tcolorbox}[
  colback=gray!8,
  colframe=gray!40,
  boxrule=0.4pt,
  arc=2pt,
  left=6pt, right=6pt, top=4pt, bottom=4pt,
  fontupper=\small\ttfamily,
  breakable,
  title={\small\bfseries Appraisal Rating}
]
\textbf{Instruction:} Imagine you are the person who wrote the following story. Read it carefully and internalize the feelings and situation described. Please indicate how much the given statement reflects how you feel about the event. When answering, please think back to the moment when the events described in the story had just ended.\\[3pt]
\textbf{My Situation:} \{scenario\}\\[3pt]
\textbf{Statement:} \{statement\}\\[3pt]
\textbf{Options:}
\begin{itemize}[nosep, leftmargin=1.5em]
  \item Strongly disagree
  \item Somewhat disagree
  \item Neither agree nor disagree
  \item Somewhat agree
  \item Strongly agree
\end{itemize}
\textbf{Constraint:} Respond only with one of the above labels. Do not provide any introductory text, explanation, or punctuation.\\[3pt]
\textbf{My Answer:}
\end{tcolorbox}

\begin{tcolorbox}[
  colback=gray!8,
  colframe=gray!40,
  boxrule=0.4pt,
  arc=2pt,
  left=6pt, right=6pt, top=4pt, bottom=4pt,
  fontupper=\small\ttfamily,
  breakable,
  title={\small\bfseries Emotion Intensity}
]
\textbf{Instruction:} Imagine you are the person who wrote the following story. Read it carefully and internalize the feelings and situation described. You have just finished experiencing these events. Answer the following reflection based on how you truly feel in that moment.\\[3pt]
\textbf{My Situation:} \{scenario\}\\[3pt]
\textbf{Question:} How \textcolor{blue}{positive} / \textcolor{red}{negative} did this situation make you feel?\\[3pt]
\textbf{Options:}
\begin{itemize}[nosep, leftmargin=1.5em]
  \item Not at all \textcolor{blue}{positive} / \textcolor{red}{negative}
  \item Very slightly \textcolor{blue}{positive} / \textcolor{red}{negative}
  \item Slightly \textcolor{blue}{positive} / \textcolor{red}{negative}
  \item Moderately \textcolor{blue}{positive} / \textcolor{red}{negative}
  \item Quite a bit \textcolor{blue}{positive} / \textcolor{red}{negative}
  \item Very \textcolor{blue}{positive} / \textcolor{red}{negative}
  \item Extremely \textcolor{blue}{positive} / \textcolor{red}{negative}
\end{itemize}
\textbf{Constraint:} Respond only with one of the above labels. Do not provide any introductory text, explanation, or punctuation.\\[3pt]
\textbf{My Answer:}
\end{tcolorbox}

\begin{tcolorbox}[
  colback=gray!8,
  colframe=gray!40,
  boxrule=0.4pt,
  arc=2pt,
  left=6pt, right=6pt, top=4pt, bottom=4pt,
  fontupper=\small\ttfamily,
  breakable,
  title={\small\bfseries Emotion Labels}
]
\textbf{Instruction:} Imagine you are the person who wrote the following story. Read it carefully and internalize the feelings and situation described. You have just finished experiencing these events. Answer the following reflection based on how you truly feel in that moment.\\[3pt]
\textbf{My Situation:} \{scenario\}\\[3pt]
\textbf{Question:} Which of the following emotion groups did you experience in this situation?\\[3pt]
\textbf{Options:} \{emotion\_labels\}\\[3pt]
\textbf{Constraint:}
\begin{enumerate}[nosep, leftmargin=1.5em]
  \item If you did not feel any \textcolor{blue}{positive} / \textcolor{red}{negative} emotions, respond only with ``None''.
  \item If you experienced any of the above, list all applicable groups exactly as they are written, separated by a semicolon (;).
  \item Do not provide any introductory text, explanation, or punctuation outside of the list.
\end{enumerate}
\textbf{My Answer:}
\end{tcolorbox}

\subsection{Supplementary Results}
\label{app:sup-tables}

\subsubsection{Supplementary Results for RQ1}
\label{app:rq1}

\paragraph{Human Evaluation of Appraisal Reasoning}
To complement automatic metrics for appraisal reasoning evaluation, we conducted a human evaluation on a sample of 100 scenarios. For each core appraisal dimension, two human evaluators were presented with the scenario, the corresponding core appraisal question, the first-person reference answer, and the reasoning responses generated by each of the six models alongside one human third-person observer response. Evaluators performed two tasks: (1) a consistency task, in which they judged independently whether each response was consistent with the first-person reference; and (2) a best-response task, in which they selected the single most consistent response from all candidates.

\begin{figure}[ht]
    \centering
    \includegraphics[width=\textwidth]{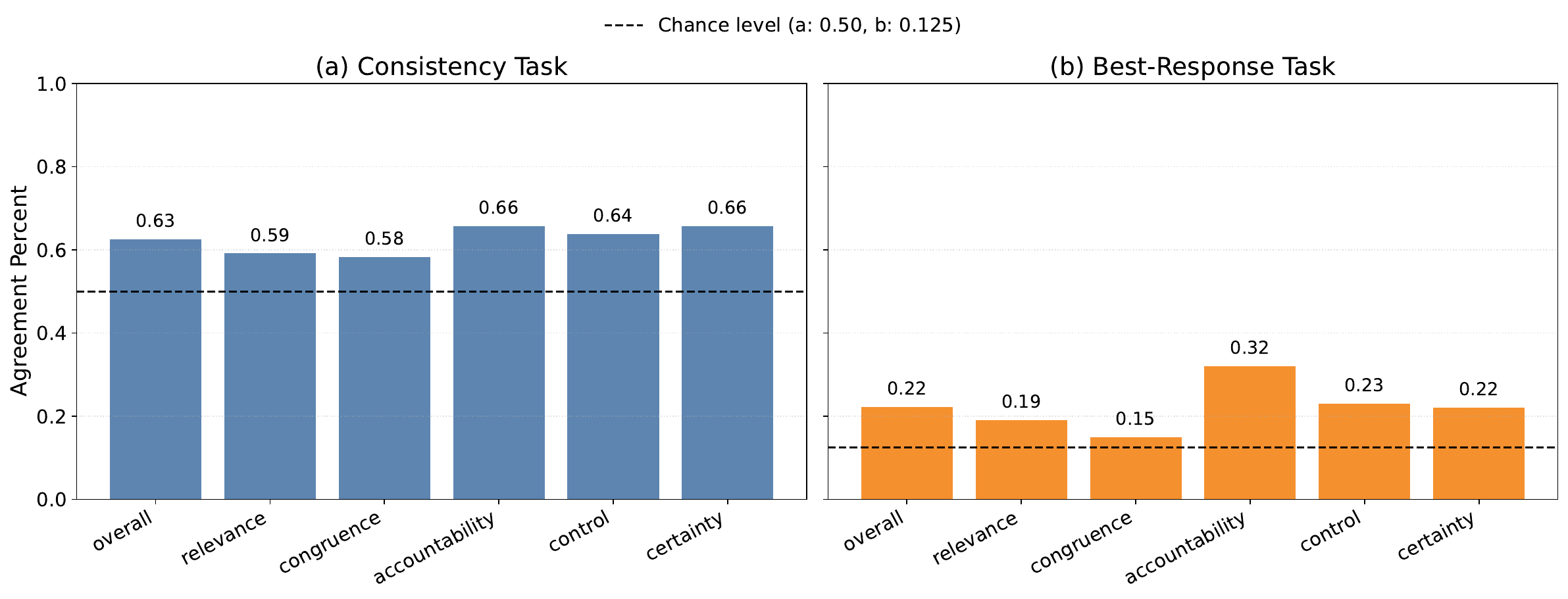}
    \caption{Inter-annotator agreement for the consistency task and best-response task, reported overall and by appraisal dimension.}
    \label{fig:human_eval_agreement}
\end{figure}

\begin{figure}[ht]
    \centering
    \includegraphics[width=\textwidth]{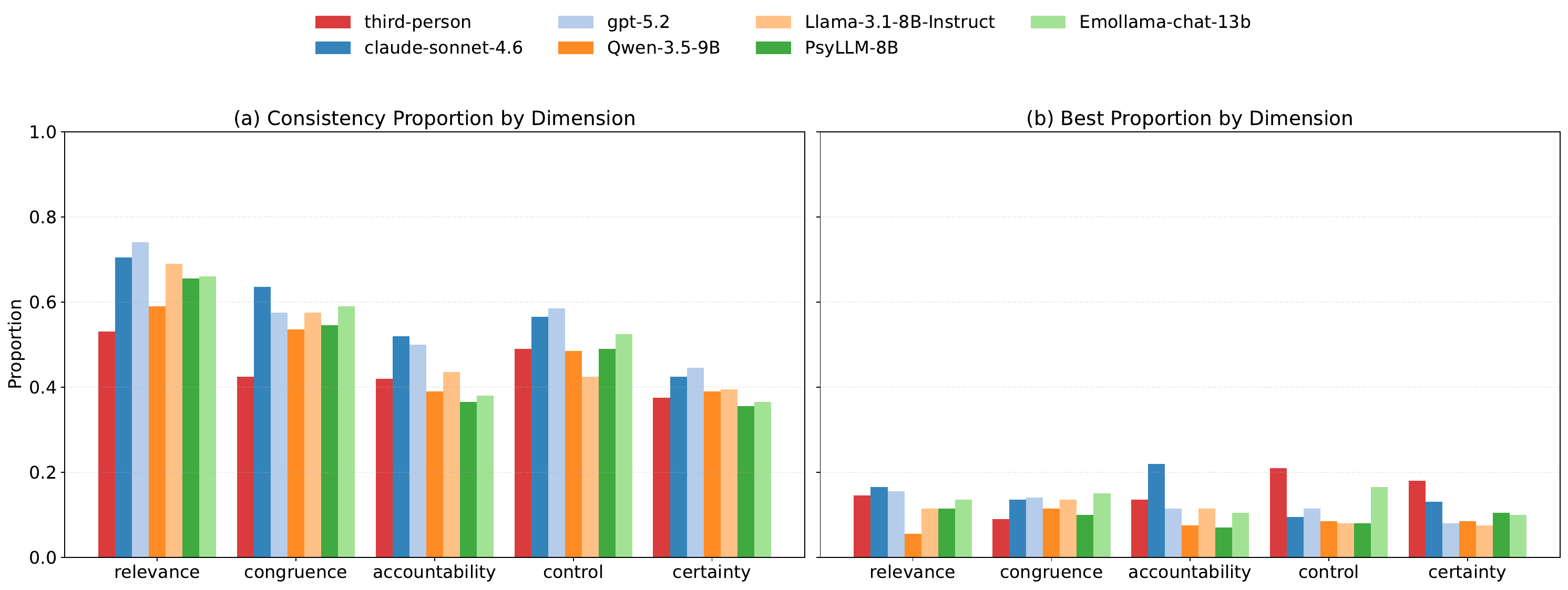}
    \caption{Consistency proportion and best-response selection rate by model and appraisal dimension.}
    \label{fig:dim_consistency_best}
\end{figure}

Inter-annotator agreement results are shown in Figure~\ref{fig:human_eval_agreement}. For the consistency task, overall Fleiss' $\kappa = 0.25$ with an agreement rate of 62.5\%, indicating fair-to-moderate reliability. For the best-response task, $\kappa = 0.10$ with an agreement rate of 22.2\%, which exceeds the random baseline of 12.5\% but reflects substantially lower consensus. Across appraisal dimensions, accountability shows the highest agreement in both tasks, while congruence shows the lowest, suggesting that judgments of goal congruence are particularly subject to individual interpretation. These results underscore the inherent difficulty of evaluating open-ended appraisal reasoning: even trained human evaluators struggle to reach consensus on what constitutes a consistent inference about another person's subjective experience.

Figure~\ref{fig:dim_consistency_best} further illustrates performance differences between models and the human observer across individual appraisal dimensions. On the consistency task, models outperform the human third-person observer across all five dimensions, with the advantage most pronounced on relevance and congruence. On the best-response selection task, however, the human observer achieves the highest proportion on control and certainty. On accountability, Claude achieves the highest best-response rate among all candidates. The following presents concrete illustrative examples.

\begin{tcolorbox}[
    colback=gray!5,
    colframe=gray!40,
    boxrule=0.5pt,
    arc=4pt,
    breakable,
    title={\textbf{Case Studies: Appraisal Reasoning}},
    fonttitle=\small
]

\small

\textbf{Example 1} \textit{(preferred: human observer; dimension: control)}

\smallskip
\textbf{Scenario:} My favorite team, Michigan, was ranked number 1 and was playing against Duke, who was ranked number 3, on Saturday. It was a big game, and I was really hoping for Michigan to win. I was watching the game, rooting for them, and I even bet on them. Unfortunately, Michigan's guards didn't have a good game, and they lost. I felt upset and frustrated after the game. Then, the new rankings came out on Monday, and now Duke is number 1 and Michigan is number 3.

\smallskip
\textbf{Question:} Did you feel like you were in control? Why?

\smallskip
\textbf{1st-person reference:} No, I didn't feel like I was in control because I was just watching the game from home.

\smallskip
\textbf{Human observer:} I had no control as I was only watching the game and I have no impact on how well they play.

\smallskip
\textbf{Claude:} No, I didn't feel in control at all. I could cheer and bet, but I couldn't influence how Michigan's guards played. Watching them struggle and losing money on the bet made me feel completely powerless and frustrated.

\tcbline

\textbf{Example 2} \textit{(preferred: human observer; dimension: certainty)}

\smallskip
\textbf{Scenario:} I accompanied my wife to the doctor to deal with a pain she's had since giving birth to our son six years ago. We had to go over the experience of that difficult birth with the doctor, and it made both of us relive it and break down in tears. We were trying to get a diagnosis for my wife's continued pain by giving as clear a description of the cause as possible. We felt emotionally drained but also pleased because we finally got a promising diagnosis after years of trying. It was important to me to ensure my wife had some clarity and hope.

\smallskip
\textbf{Question:} Did things go as you expected? If not, what was unexpected?

\smallskip
\textbf{1st-person reference:} Things went pretty much as expected. After 6 years of trying I suppose it was quite surprising to get a diagnosis but I was optimistic.

\smallskip
\textbf{Human observer:} It went good as we got a promising diagnosis, that is great news.

\smallskip
\textbf{Claude:} I expected it to be difficult, but I didn't anticipate how deeply reliving that birth would affect us both. Breaking down in tears caught me off guard. What I didn't expect was leaving with actual hope---a real diagnosis after years of nothing. That was a relief I hadn't let myself count on.

\tcbline

\textbf{Example 3} \textit{(preferred: Claude; dimension: relevance)}

\smallskip
\textbf{Scenario:} Me and my family had booked a trip to Disneyland Paris, flying from Edinburgh to Paris with Air France. We were all set for a fun family holiday, with only two days planned at the theme park. When we got to the airport, we boarded the plane and were just about to take off when the flight was cancelled, completely ruining our trip. I felt very, very angry and heartbroken at the same time. It was especially upsetting for our kids because we had surprised them with the trip that morning, and they were so excited, only to be told they couldn't go.

\smallskip
\textbf{Question:} When the event happened, how much did it matter to you? Why?

\smallskip
\textbf{1st-person reference:} It happened last weekend. It was our first trip there so it mattered a lot! Creating memories with our kids that we couldn't then do.

\smallskip
\textbf{Human observer:} It mattered a lot because it was a massive disappointment to my kids.

\smallskip
\textbf{Claude:} It mattered enormously. We'd planned this special surprise for our kids, and seeing their excitement crushed was devastating. It wasn't just a holiday---it was a magical moment stolen from them. The anger and heartbreak I felt was overwhelming; I'd never felt so helpless.

\tcbline

\textbf{Example 4} \textit{(preferred: Claude; dimension: congruence)}

\smallskip
\textbf{Scenario:} A friend's mother passed away this past weekend. Even though it was expected---she was 96 and had been in ill-health and under hospice care for some time---it's always emotional to lose someone who has been part of your life. I wanted to help and be there for my friend and their family. Unfortunately, my friend had the flu and didn't want anyone to come over. I didn't really need anything other than to support my friend, which I did as much as possible from a distance. I suppose it left me feeling `OK'. At my age and with my experience, death is becoming common and I know how to deal with it. No, other than we will miss the mother.

\smallskip
\textbf{Question} In what ways did this situation help or hurt you?

\smallskip
\textbf{1st-person reference:} The situation didn't really help or hurt me.

\smallskip
\textbf{Human observer:} It hurt me as someone who I knew died.

\smallskip
\textbf{Claude:} It didn't really hurt me personally. I know how to handle loss at my stage of life. The only frustration was not being able to be there in person for my friend. Supporting from a distance felt limiting, but I managed okay.

\tcbline

\textbf{Example 5} \textit{(preferred: Claude; dimension: accountability)}

\smallskip
\textbf{Scenario:} It was a very sad time because my friend had to have her dear, precious dog put to sleep. I've been giving my friend lots of support and care, and I really wish I could have been able to see her more during this tough time. I'm feeling very sad myself because he was a really lovely, friendly dog and a real miss for both my dear friend and me. You feel low and empty when something like this happens. It's important to keep positive for my friend, so I've been telling her about all the wonderful memories we have and what a great life he had with her.

\smallskip
\textbf{Question:} Who or what did you feel was most responsible for all of this?

\smallskip
\textbf{1st-person reference:} No one is responsible as nature and age is sadly what we all face, as he was 14 years old.

\smallskip
\textbf{Human observer:} It's just life.

\smallskip
\textbf{Claude:} Honestly, no one was really responsible---it was simply the heartbreak of losing a beloved pet to old age or illness. Nature takes its course, and that's the hardest, most unavoidable truth. No blame, just deep sadness.

\end{tcolorbox}

\paragraph{Fine-Grained Appraisal and Emotion Results} 

Tables~\ref{tab:rq1_reasoning_1} and \ref{tab:rq1_reasoning_2} report per-dimension appraisal reasoning results (BLEU, ROUGE-L, BERTScore) across the five core appraisal dimensions. Table~\ref{tab:rq1_appraisal} reports normalized RMSE for each of the 22 appraisal rating sub-dimensions. Tables~\ref{tab:rq1_neglabel} and \ref{tab:rq1_poslabel} report per-label precision, recall, and F1 for negative and positive emotion categories, respectively.

\begin{table*}[ht]
\centering
\footnotesize
\setlength{\tabcolsep}{4pt}
\caption{Per-dimension appraisal reasoning results (Part 1: Relevance, Congruence, Accountability). Bold indicates the best, and underlining indicates the second best.}
\label{tab:rq1_reasoning_1}
\begin{tabular}{l ccc ccc ccc}
\toprule
& \multicolumn{3}{c}{\textbf{Relevance}} 
& \multicolumn{3}{c}{\textbf{Congruence}} 
& \multicolumn{3}{c}{\textbf{Accountability}} \\
\cmidrule(lr){2-4} \cmidrule(lr){5-7} \cmidrule(lr){8-10}
Model 
& BLEU & ROUGE-L & BERTScore 
& BLEU & ROUGE-L & BERTScore 
& BLEU & ROUGE-L & BERTScore \\
\midrule
Human    & 0.012 & 0.186 & 0.879 & 0.005 & 0.163 & 0.879 & 0.004 & 0.144 & 0.871 \\
\midrule
Claude   & 0.012 & 0.175 & 0.871 & 0.003 & 0.153 & 0.875 & 0.004 & 0.138& 0.867\\
GPT      & \underline{0.018} & \textbf{0.182} & 0.870 & 0.003 & \underline{0.155} & 0.871 & \underline{0.005} & \underline{0.145}& 0.867\\
Qwen     & 0.001 & 0.147 & \textbf{0.874} & 0.000 & 0.146 & \underline{0.878} & 0.002 & 0.134 & \underline{0.870}\\
Llama    & \textbf{0.020} & 0.174 & 0.870 & \underline{0.004} & 0.133 & 0.866 & \underline{0.005} & 0.119 & 0.860 \\
PsyLLM   & 0.003 & 0.149 & 0.864 & 0.003 & 0.137 & 0.868 & 0.001 & 0.110 & 0.852 \\
EmoLlama & 0.014 & \underline{0.181} & \underline{0.873} & \textbf{0.005} & \textbf{0.162} & \textbf{0.879} & \textbf{0.009} & \textbf{0.160} & \textbf{0.874} \\
\bottomrule
\end{tabular}
\end{table*}

\begin{table*}[ht]
\centering
\footnotesize
\caption{Per-dimension appraisal reasoning results (Part 2: Control, Certainty). Bold indicates the best, and underlining indicates the second best.}
\label{tab:rq1_reasoning_2}
\begin{tabular}{l ccc ccc}
\toprule
& \multicolumn{3}{c}{\textbf{Control}} 
& \multicolumn{3}{c}{\textbf{Certainty}} \\
\cmidrule(lr){2-4} \cmidrule(lr){5-7}
Model 
& BLEU & ROUGE-L & BERTScore 
& BLEU & ROUGE-L & BERTScore \\
\midrule
Human    & 0.009 & 0.188 & 0.877 & 0.004 & 0.135 & 0.870 \\
\midrule
Claude   & 0.009 & \underline{0.180} & 0.876& 0.007 & \underline{0.139} & 0.869\\
GPT      & 0.007 & 0.167 & 0.872 & 0.005 & 0.132 & 0.864 \\
Qwen     & 0.003 & 0.165 & \underline{0.879} & 0.004 & 0.130 & \underline{0.872} \\
Llama    & \underline{0.011} & 0.148 & 0.868 & 0.007 & 0.121 & 0.864 \\
PsyLLM   & 0.009 & 0.160 & 0.869 & \underline{0.010} & 0.131 & 0.864 \\
EmoLlama & \textbf{0.023} & \textbf{0.208} & \textbf{0.884} & \textbf{0.013} & \textbf{0.162} & \textbf{0.880} \\
\bottomrule
\end{tabular}
\end{table*}

\begin{table*}[ht]
\centering
\small
\caption{Per-dimension appraisal rating results (normalized RMSE, lower is better). Best result among LLMs in each row is in \textbf{bold}. \colorbox{worse}{Red} cells indicate worse performance than Third-person.}
\label{tab:rq1_appraisal}
\begin{tabular}{l ccccccc}
\toprule
\textbf{Dimension} & \textbf{Third-person} & \textbf{Claude} & \textbf{GPT} & \textbf{Qwen} & \textbf{Llama} & \textbf{PsyLLM} & \textbf{EmoLlama} \\
\midrule
\multicolumn{8}{l}{\textit{Relevance}} \\
\quad General           & 0.217 & \textbf{0.182} & 0.180 & 0.186 & \textbf{0.176} & 0.181 & 0.195 \\
\quad Urgency           & 0.417 & 0.406 & 0.417 & \cellcolor{worse}0.477 & \cellcolor{worse}0.449 & \textbf{0.398} & \cellcolor{worse}0.501 \\
\quad Goals             & 0.442 & \cellcolor{worse}0.500 & 0.435 & \cellcolor{worse}0.522 & \cellcolor{worse}0.444 & \textbf{0.383} & \cellcolor{worse}0.458 \\
\quad Bodily motives    & 0.441 & \textbf{0.367} & 0.380 & \cellcolor{worse}0.451 & 0.436 & 0.384 & \cellcolor{worse}0.515 \\
\quad Social motives    & 0.331 & \textbf{0.283} & 0.287 & 0.304 & 0.298 & 0.285 & 0.313 \\
\quad Identity motives  & 0.420 & 0.356 & 0.365 & 0.414 & \textbf{0.353} & 0.360 & 0.381 \\
\midrule
\multicolumn{8}{l}{\textit{Certainty}} \\
\quad Construal         & 0.361 & 0.325 & 0.324 & 0.355 & \textbf{0.274} & 0.305 & 0.281 \\
\quad Outlook           & 0.407 & 0.396 & 0.404 & \cellcolor{worse}0.538 & \textbf{0.380} & \textbf{0.380} & 0.406 \\
\quad Predictability    & 0.414 & \textbf{0.352} & 0.390 & \cellcolor{worse}0.497 & \cellcolor{worse}0.475 & 0.384 & \cellcolor{worse}0.486 \\
\quad Novelty           & 0.431 & \textbf{0.371} & 0.380 & \cellcolor{worse}0.452 & 0.409 & 0.400 & 0.422 \\
\midrule
\multicolumn{8}{l}{\textit{Congruence}} \\
\quad General                   & 0.315 & 0.275 & \textbf{0.262} & 0.306 & \cellcolor{worse}0.344 & 0.279 & \cellcolor{worse}0.400 \\
\quad Outlook                   & 0.373 & 0.333 & 0.334 & \cellcolor{worse}0.377 & 0.333 & \textbf{0.312} & 0.344 \\
\quad Positive prediction error & 0.349 & 0.313 & 0.306 & \cellcolor{worse}0.361 & \cellcolor{worse}0.377 & \textbf{0.298} & \cellcolor{worse}0.500 \\
\quad Negative prediction error & 0.383 & 0.341 & \textbf{0.327} & \cellcolor{worse}0.383 & 0.344 & \textbf{0.327} & \cellcolor{worse}0.482 \\
\midrule
\multicolumn{8}{l}{\textit{Control}} \\
\quad General   & 0.409 & 0.353 & 0.352 & 0.400 & 0.383 & \textbf{0.332} & \cellcolor{worse}0.494 \\
\quad Select    & 0.443 & \cellcolor{worse}0.462 & 0.409 & \cellcolor{worse}0.506 & 0.441 & \textbf{0.398} & \cellcolor{worse}0.446 \\
\quad Vicarious & 0.407 & 0.373 & \textbf{0.368} & \cellcolor{worse}0.418 & \cellcolor{worse}0.567 & 0.407& \cellcolor{worse}0.591 \\
\quad Effortful & 0.372 & 0.359 & \cellcolor{worse}0.412 & \cellcolor{worse}0.501 & 0.339 & \textbf{0.337} & 0.360 \\
\midrule
\multicolumn{8}{l}{\textit{Accountability}} \\
\quad Self           & 0.389 & \textbf{0.323} & 0.341 & \cellcolor{worse}0.447 & \cellcolor{worse}0.462 & 0.359 & \cellcolor{worse}0.510 \\
\quad Other          & 0.402 & \textbf{0.393} & \cellcolor{worse}0.451 & \cellcolor{worse}0.549 & \cellcolor{worse}0.454 & 0.385 & \cellcolor{worse}0.450 \\
\quad Intentionality & 0.405 & \textbf{0.332} & \cellcolor{worse}0.431 & \cellcolor{worse}0.468 & \cellcolor{worse}0.427 & 0.377 & \cellcolor{worse}0.511 \\
\quad Fairness       & 0.355 & 0.344 & 0.348 & \cellcolor{worse}0.360 & 0.351 & \textbf{0.325} & \cellcolor{worse}0.500 \\
\midrule
Macro-averaged & 0.389 & 0.357 & 0.365 & \cellcolor{worse}0.430 & \cellcolor{worse}0.395 & \textbf{0.349} & \cellcolor{worse}0.443 \\
\bottomrule
\end{tabular}
\end{table*}

\begin{table*}[ht]
\centering
\small
\caption{Per-label negative emotion results. Best P, R, and F1 among LLMs in each column are in \textbf{bold}. \colorbox{worse}{Red} F1 cells indicate worse performance than Third-person.}
\label{tab:rq1_neglabel}
\resizebox{\textwidth}{!}{%
\begin{tabular}{l l cccccccccc c}
\toprule
& & {\textbf{Angry}}
& {\textbf{Worried}}
& {\textbf{Sad}}
& {\textbf{Disgust}}
& {\textbf{Despair}}
& {\textbf{Ashamed}}
& {\textbf{Lonely}}
& {\textbf{Panicked}}
& {\textbf{Guilty}}
& {\textbf{Confused}}
& {\textbf{Macro-avg}} \\
\midrule
\multirow{3}{*}{Third-person}
& P  & 0.737 & 0.675 & 0.777 & 0.321 & 0.503 & 0.319 & 0.370 & 0.368 & 0.213 & 0.313 & 0.459 \\
& R  & 0.651 & 0.596 & 0.677 & 0.316 & 0.437 & 0.324 & 0.382 & 0.382 & 0.311 & 0.351 & 0.443 \\
& F1 & 0.691 & 0.633 & 0.723 & 0.319 & 0.468 & 0.321 & 0.376 & 0.375 & 0.253 & 0.331 & 0.449 \\
\midrule
\multirow{3}{*}{Claude}
& P  & 0.885 & \textbf{0.741} & 0.813 & 0.432 & \textbf{0.657} & \textbf{0.639} & 0.474 & \textbf{0.602} & 0.376 & \textbf{0.411} & \textbf{0.603} \\
& R  & 0.561 & 0.726 & 0.808 & 0.168 & 0.461 & \textbf{0.310} & 0.517 & \textbf{0.384} & 0.409 & 0.299 & 0.464 \\
& F1 & 0.686 & 0.733 & \textbf{0.811} & \cellcolor{worse}0.242 & 0.542 & \textbf{0.417} & \textbf{0.495} & \textbf{0.469} & \textbf{0.392} & 0.346 & \textbf{0.513} \\
\midrule
\multirow{3}{*}{GPT}
& P  & 0.873 & 0.721 & \textbf{0.818} & \textbf{0.588} & 0.613 & 0.547 & \textbf{0.481} & 0.558 & 0.434 & 0.372 & 0.601 \\
& R  & \textbf{0.578} & 0.786 & 0.747 & 0.105 & 0.510 & 0.325 & 0.432 & 0.362 & 0.355 & 0.318 & 0.452 \\
& F1 & \textbf{0.696} & \textbf{0.752} & 0.781 & \cellcolor{worse}0.179 & 0.557 & 0.408 & 0.455 & 0.439 & 0.391 & 0.343 & 0.500 \\
\midrule
\multirow{3}{*}{Qwen}
& P  & 0.896 & 0.724 & 0.807 & 0.366 & 0.566 & 0.525 & 0.425 & 0.580 & 0.286 & 0.376 & 0.555 \\
& R  & 0.543 & 0.750 & 0.745 & 0.179 & \textbf{0.649} & 0.305 & \textbf{0.595} & 0.364 & \textbf{0.531} & \textbf{0.378} & \textbf{0.504} \\
& F1 & 0.677 & 0.737 & 0.775 & \cellcolor{worse}0.240 & \textbf{0.605} & 0.386 & \textbf{0.496} & 0.447 & 0.371 & \textbf{0.377} & 0.511 \\
\midrule
\multirow{3}{*}{Llama}
& P  & 0.872 & 0.669 & 0.799 & 0.435 & 0.553 & 0.500 & 0.423 & 0.494 & 0.273 & 0.349 & 0.537 \\
& R  & 0.506 & 0.749 & 0.748 & 0.160 & 0.567 & 0.287 & 0.330 & 0.381 & 0.385 & 0.123 & 0.424 \\
& F1 & \cellcolor{worse}0.640 & 0.707 & 0.773 & \cellcolor{worse}0.234 & 0.560 & 0.365 & \cellcolor{worse}0.371 & 0.430 & 0.319 & \cellcolor{worse}0.182 & 0.458 \\
\midrule
\multirow{3}{*}{PsyLLM}
& P  & 0.918 & 0.708 & 0.786 & 0.372 & 0.544 & 0.467 & 0.354 & 0.570 & 0.246 & 0.340 & 0.531 \\
& R  & 0.481 & 0.638 & 0.811 & \textbf{0.229} & 0.534 & 0.213 & 0.541 & 0.288 & 0.394 & 0.291 & 0.442 \\
& F1 & \cellcolor{worse}0.631 & 0.672 & 0.798 & \cellcolor{worse}0.283 & 0.539 & \cellcolor{worse}0.292 & 0.428 & 0.382 & 0.303 & \cellcolor{worse}0.313 & 0.464 \\
\midrule
\multirow{3}{*}{EmoLlama}
& P  & \textbf{0.953} & 0.642 & 0.697 & 0.187 & 0.610 & 0.467 & 0.435 & 0.342 & \textbf{0.533} & 0.260 & 0.513 \\
& R  & 0.397 & \textbf{0.791} & \textbf{0.812} & 0.093 & 0.169 & 0.216 & 0.232 & 0.057 & 0.153 & 0.067 & 0.299 \\
& F1 & \cellcolor{worse}0.560 & 0.709 & 0.750 & \cellcolor{worse}0.123 & \cellcolor{worse}0.265 & \cellcolor{worse}0.295 & \cellcolor{worse}0.302 & \cellcolor{worse}0.098 & 0.237 & \cellcolor{worse}0.106 & \cellcolor{worse}0.345 \\
\bottomrule
\end{tabular}%
}
\end{table*}

\begin{table*}[ht]
\centering
\small
\caption{Per-label positive emotion results. Best P, R, and F1 among LLMs in each column are in \textbf{bold}. \colorbox{worse}{Red} F1 cells indicate worse performance than Third-person.}
\label{tab:rq1_poslabel}
\resizebox{\textwidth}{!}{%
\begin{tabular}{l l cccccccccc c}
\toprule
& & \textbf{Hopeful}
& \textbf{Grateful}
& \textbf{Glad}
& \textbf{Love}
& \textbf{Amused}
& \textbf{Sympathy}
& \textbf{Inspiration}
& \textbf{Calm}
& \textbf{Proud}
& \textbf{Excited}
& \textbf{Macro-avg} \\
\midrule
\multirow{3}{*}{Third-person}
& P  & 0.581 & 0.601 & 0.662 & 0.563 & 0.276 & 0.214 & 0.392 & 0.328 & 0.573 & 0.563 & 0.475 \\
& R  & 0.603 & 0.588 & 0.677 & 0.554 & 0.223 & 0.281 & 0.397 & 0.340 & 0.557 & 0.451 & 0.467 \\
& F1 & 0.592 & 0.594 & 0.670 & 0.558 & 0.247 & 0.243 & 0.395 & 0.334 & 0.565 & 0.501 & 0.470 \\
\midrule
\multirow{3}{*}{Claude}
& P  & 0.506 & 0.602 & 0.684 & 0.451 & \textbf{0.750} & 0.206 & 0.526 & \textbf{0.345} & 0.530 & 0.671 & 0.527 \\
& R  & \textbf{0.425} & \textbf{0.596} & \textbf{0.849} & 0.773 & \textbf{0.188} & 0.586 & 0.276 & \textbf{0.484} & \textbf{0.671} & \textbf{0.605} & \textbf{0.545} \\
& F1 & \cellcolor{worse}0.462 & 0.599 & 0.757 & 0.569 & \textbf{0.300} & 0.305 & \cellcolor{worse}0.362 & \textbf{0.403} & 0.592 & 0.636 & \textbf{0.499} \\
\midrule
\multirow{3}{*}{GPT}
& P  & 0.571 & 0.607 & 0.696 & 0.475 & 0.667 & \textbf{0.218} & \textbf{0.574} & 0.317 & 0.580 & 0.695 & \textbf{0.540} \\
& R  & 0.317 & 0.472 & 0.841 & 0.778 & \textbf{0.188} & 0.495 & 0.186 & 0.418 & 0.567 & \textbf{0.605} & 0.487 \\
& F1 & \cellcolor{worse}0.408 & \cellcolor{worse}0.531 & 0.761 & \textbf{0.590} & 0.293 & 0.303 & \cellcolor{worse}0.281 & \cellcolor{worse}0.360 & 0.573 & \textbf{0.647} & 0.475 \\
\midrule
\multirow{3}{*}{Qwen}
& P  & 0.632 & 0.597 & 0.714 & 0.389 & 0.705 & 0.207 & 0.393 & 0.317 & 0.575 & 0.649 & 0.518 \\
& R  & 0.365 & 0.469 & 0.822 & 0.734 & 0.109 & 0.598 & \textbf{0.374} & 0.272 & 0.636 & 0.595 & 0.497 \\
& F1 & \cellcolor{worse}\textbf{0.463} & \cellcolor{worse}0.525 & \textbf{0.764} & \cellcolor{worse}0.508 & \cellcolor{worse}0.189 & \textbf{0.307} & \textbf{0.383} & \cellcolor{worse}0.293 & \textbf{0.604} & \cellcolor{worse}0.621 & \cellcolor{worse}0.466 \\
\midrule
\multirow{3}{*}{Llama}
& P  & 0.702 & 0.587 & 0.673 & 0.471 & 0.511 & 0.166 & 0.483 & 0.200 & 0.570 & 0.601 & 0.496 \\
& R  & 0.212 & 0.423 & 0.838 & 0.672 & 0.062 & 0.521 & 0.159 & 0.177 & 0.570 & 0.515 & 0.415 \\
& F1 & \cellcolor{worse}0.326 & \cellcolor{worse}0.492 & 0.746 & \cellcolor{worse}0.554 & \cellcolor{worse}0.111 & 0.251 & \cellcolor{worse}0.239 & \cellcolor{worse}0.188 & 0.570 & 0.554 & \cellcolor{worse}0.403 \\
\midrule
\multirow{3}{*}{PsyLLM}
& P  & 0.606 & 0.566 & 0.626 & 0.418 & 0.057 & 0.149 & 0.470 & 0.207 & 0.505 & 0.629 & 0.423 \\
& R  & 0.339 & 0.347 & 0.828 & \textbf{0.800} & 0.022 & 0.542 & 0.223 & 0.236 & 0.615 & 0.380 & 0.433 \\
& F1 & \cellcolor{worse}0.434 & \cellcolor{worse}0.430 & 0.713 & \cellcolor{worse}0.549 & \cellcolor{worse}0.031 & \cellcolor{worse}0.233 & \cellcolor{worse}0.303 & \cellcolor{worse}0.220 & \cellcolor{worse}0.554 & \cellcolor{worse}0.474 & \cellcolor{worse}0.394 \\
\midrule
\multirow{3}{*}{EmoLlama}
& P  & \textbf{0.755} & \textbf{0.697} & \textbf{0.761} & \textbf{0.601} & 0.000 & 0.119 & 0.423 & 0.191 & \textbf{0.598} & \textbf{0.838} & 0.498 \\
& R  & 0.206 & 0.569 & 0.596 & 0.539 & 0.000 & \textbf{0.733} & 0.087 & 0.148 & 0.301 & 0.064 & 0.324 \\
& F1 & \cellcolor{worse}0.324 & \textbf{0.626} & \cellcolor{worse}0.668 & 0.569 & \cellcolor{worse}0.000 & \cellcolor{worse}0.205 & \cellcolor{worse}0.144 & \cellcolor{worse}0.166 & \cellcolor{worse}0.400 & \cellcolor{worse}0.119 & \cellcolor{worse}0.322 \\
\bottomrule
\end{tabular}%
}
\end{table*}

\subsubsection{Supplementary Results for RQ2}
\label{app:rq2}

\begin{figure*}[ht]
\centering
\includegraphics[width=\textwidth]{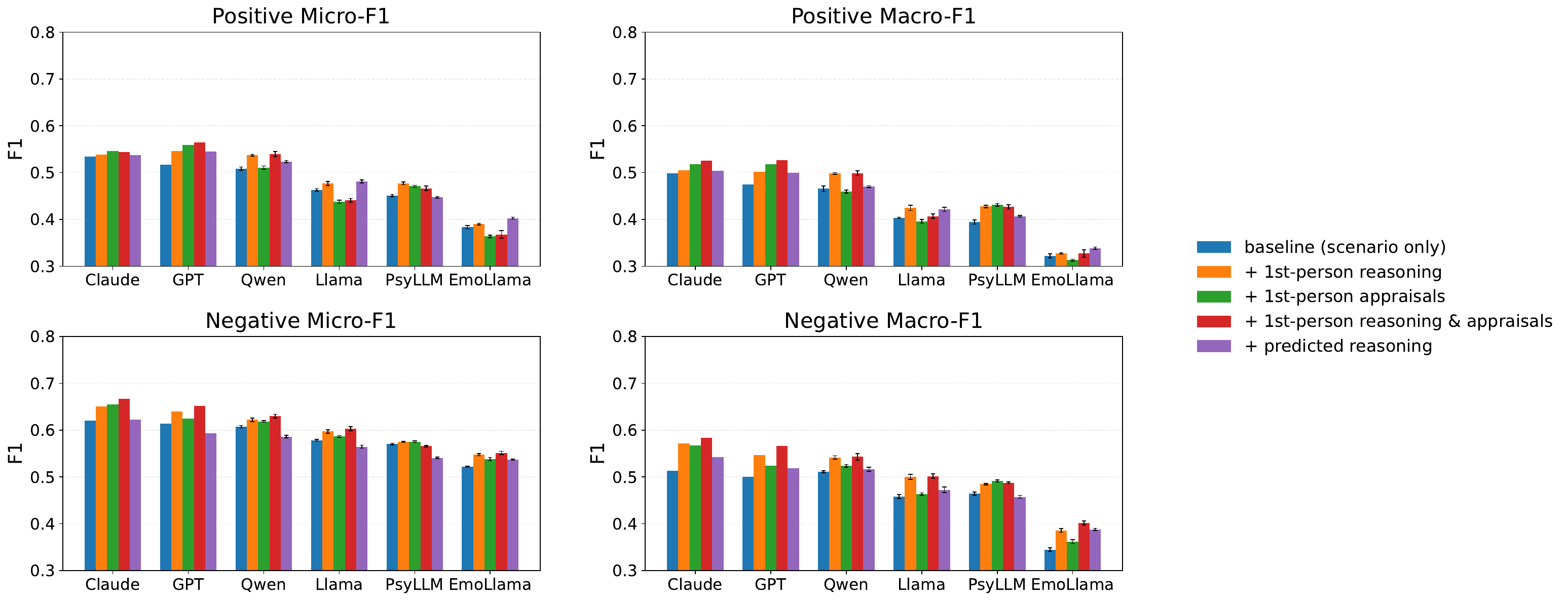}
\caption{Comparison of positive and negative emotion Micro-F1 and Macro-F1 performance across models and different input settings.}
\label{fig:appendix_emotion_f1}
\end{figure*}

\begin{figure*}[ht]
    \centering
    \includegraphics[width=\textwidth]{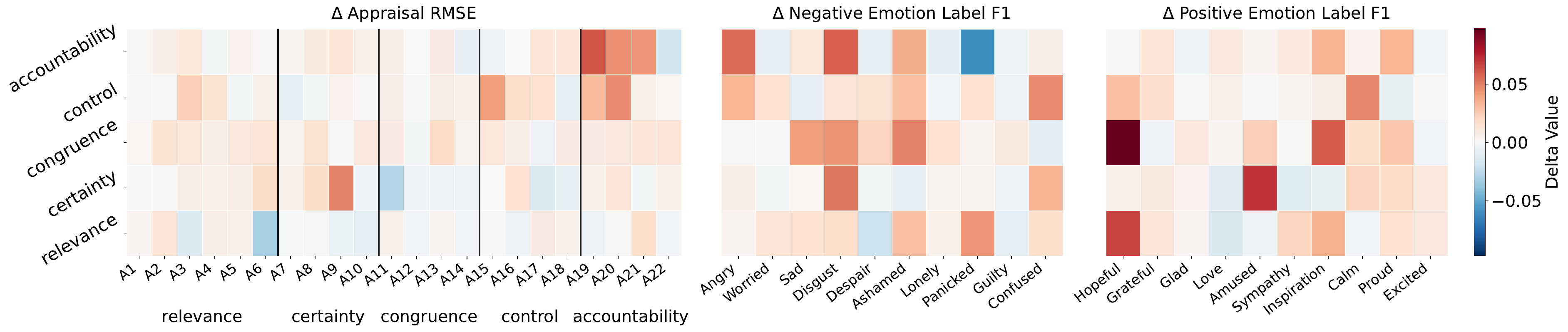}
    \caption{Heatmap of performance change ($\Delta$) from injecting dimension-specific appraisal reasoning into the scenario input (Qwen). Red indicates improvement, blue indicates degradation.}
    \label{fig:rq2_heatmap_qwen}
\end{figure*}

\begin{figure*}[ht]
    \centering
    \includegraphics[width=\textwidth]{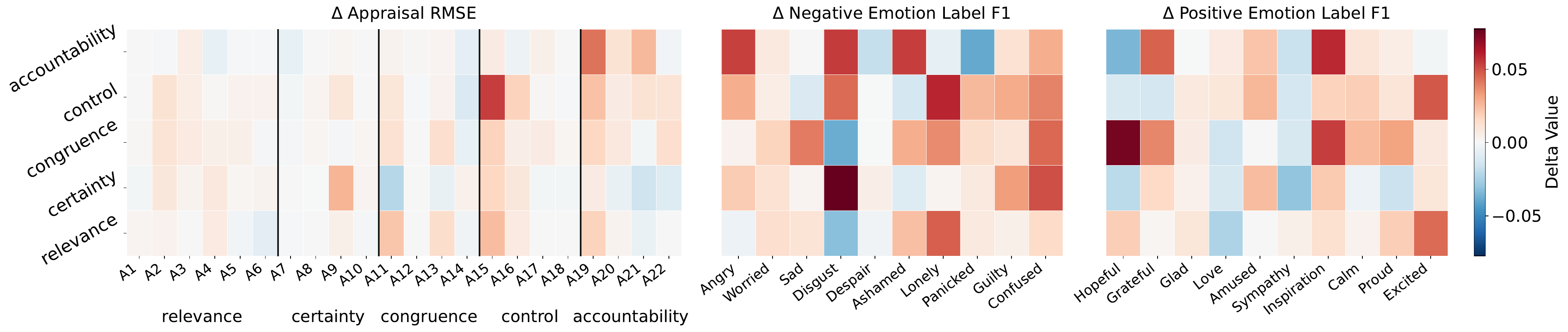}
    \caption{Heatmap of performance change ($\Delta$) from injecting dimension-specific appraisal reasoning into the scenario input (Llama). Red indicates improvement, blue indicates degradation.}
    \label{fig:rq2_heatmap_llama}
\end{figure*}

\begin{figure*}[ht]
    \centering
    \includegraphics[width=\textwidth]{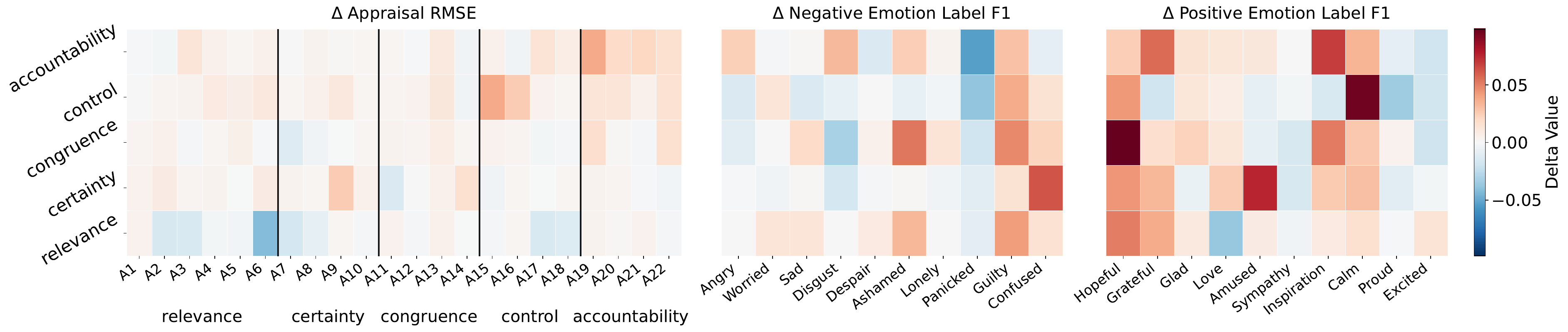}
    \caption{Heatmap of performance change ($\Delta$) from injecting dimension-specific appraisal reasoning into the scenario input (PsyLLM). Red indicates improvement, blue indicates degradation.}
    \label{fig:rq2_heatmap_psyllm}
\end{figure*}

\begin{figure*}[ht]
    \centering
    \includegraphics[width=\textwidth]{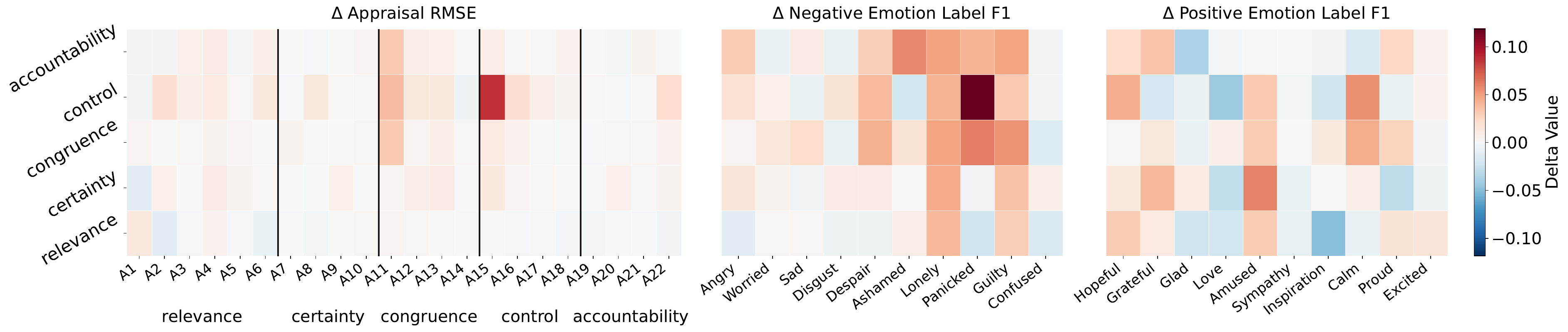}
    \caption{Heatmap of performance change ($\Delta$) from injecting dimension-specific appraisal reasoning into the scenario input (EmoLlama). Red indicates improvement, blue indicates degradation.}
    \label{fig:rq2_heatmap_emollama}
\end{figure*}

Figure~\ref{fig:appendix_emotion_f1} compares positive and negative emotion Micro-F1 and Macro-F1 across models and input settings. Figures~\ref{fig:rq2_heatmap_qwen}-\ref{fig:rq2_heatmap_emollama} present heatmaps of performance change ($\Delta$) from injecting dimension-specific appraisal reasoning for each open-source model (Qwen, Llama, PsyLLM, and EmoLlama). Dimension-specific appraisal reasoning injection experiments were conducted on open-source models only, as the API costs of GPT-5.2 and Claude-Sonnet-4.6 made exhaustive per-dimension evaluation impractical.

\subsubsection{Supplementary Results for RQ3}
\label{app:rq3}

\begin{figure*}[ht]
    \centering
    \includegraphics[width=\textwidth]{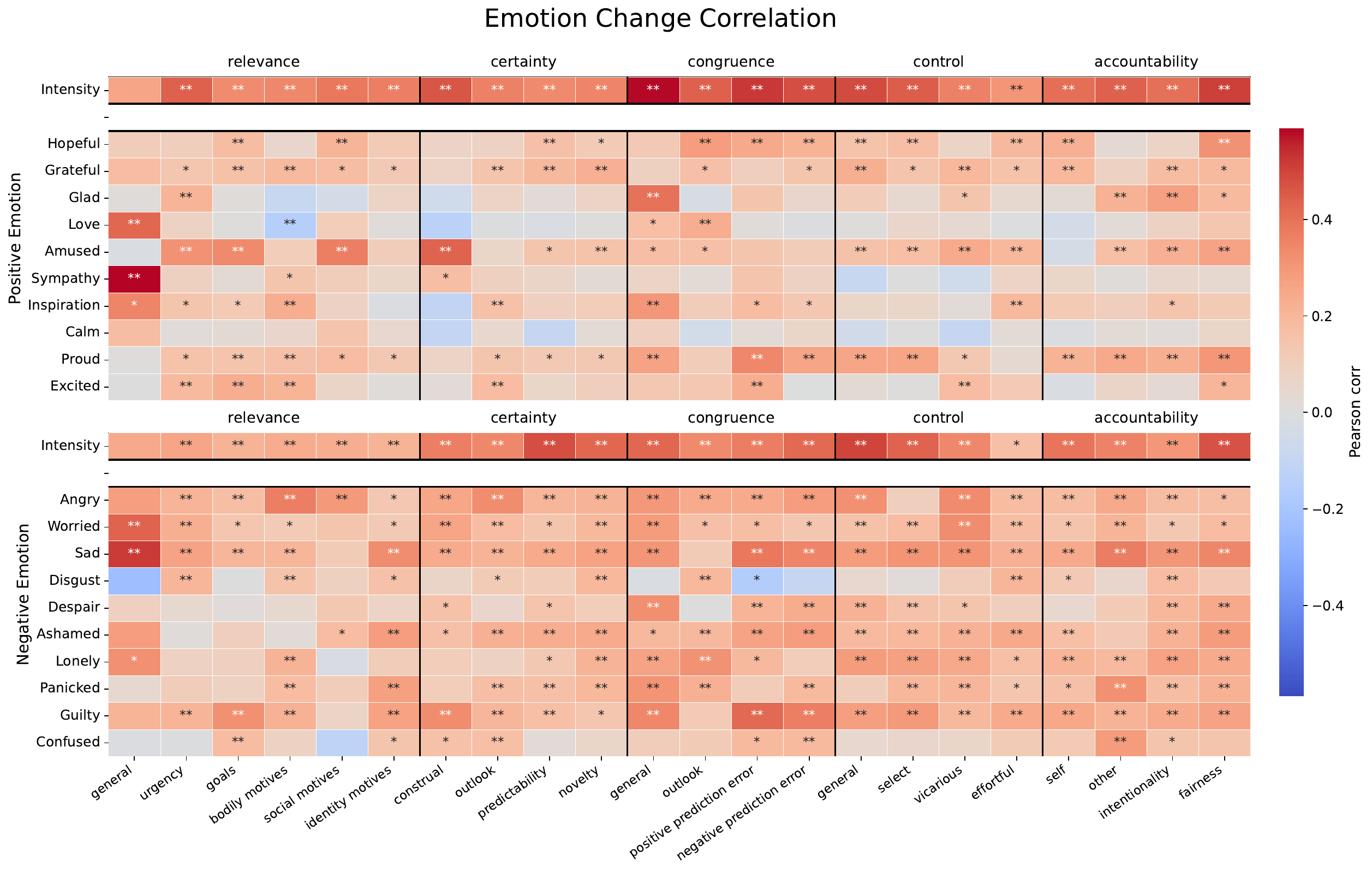}
    \caption{Pearson correlation between model-predicted emotion change and human annotation change after substituting first-person appraisal reasoning with third-person reasoning (Claude). Red indicates positive correlation; $*p < .05$, $**p < .01$.}
    \label{fig:rq3_claude}
\end{figure*}

\begin{figure*}[ht]
    \centering
    \includegraphics[width=\textwidth]{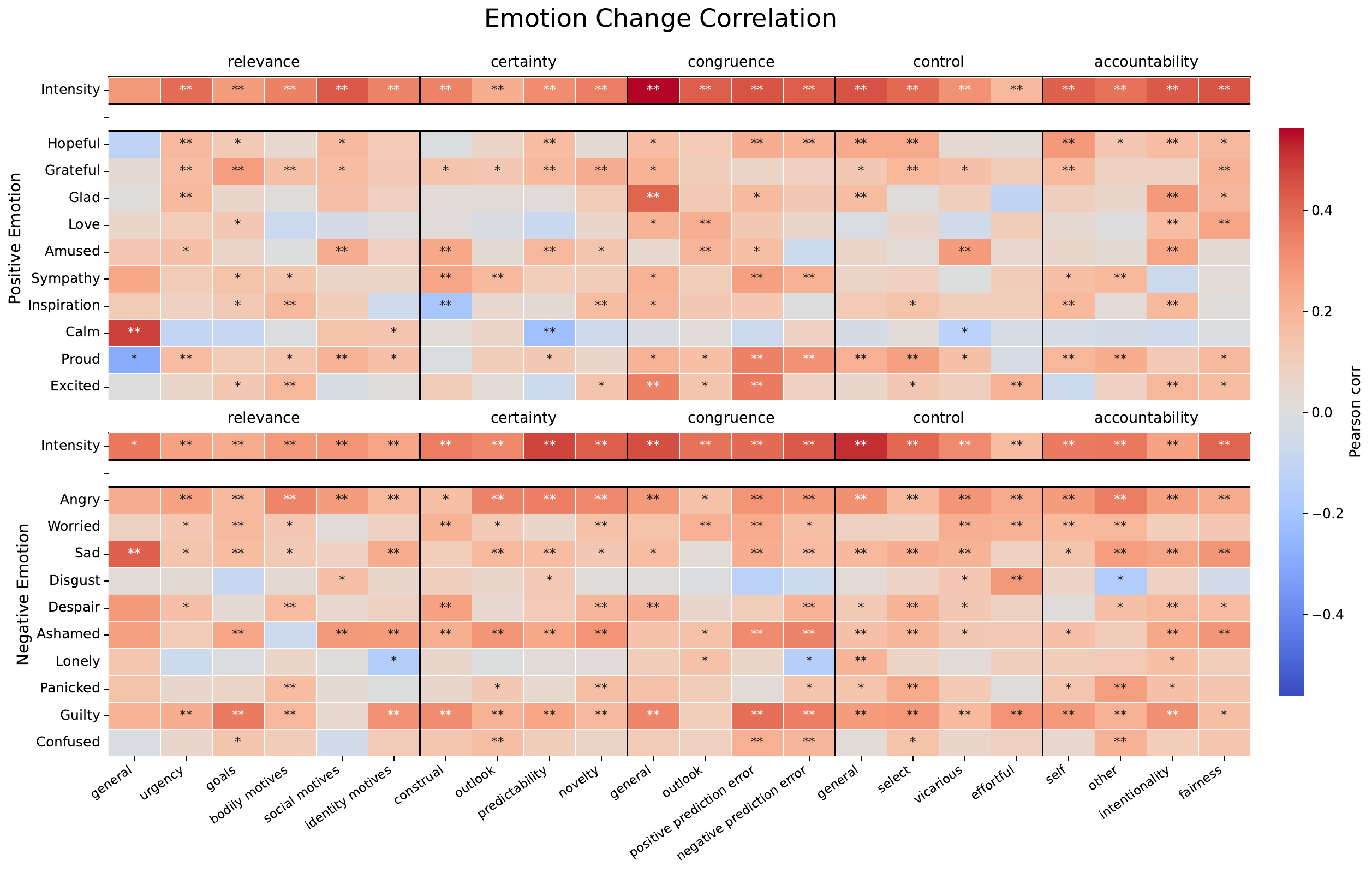}
    \caption{Pearson correlation between model-predicted emotion change and human annotation change after substituting first-person appraisal reasoning with third-person reasoning (GPT). Red indicates positive correlation; $*p < .05$, $**p < .01$.}
    \label{fig:rq3_gpt}
\end{figure*}

\begin{figure*}[ht]
    \centering
    \includegraphics[width=\textwidth]{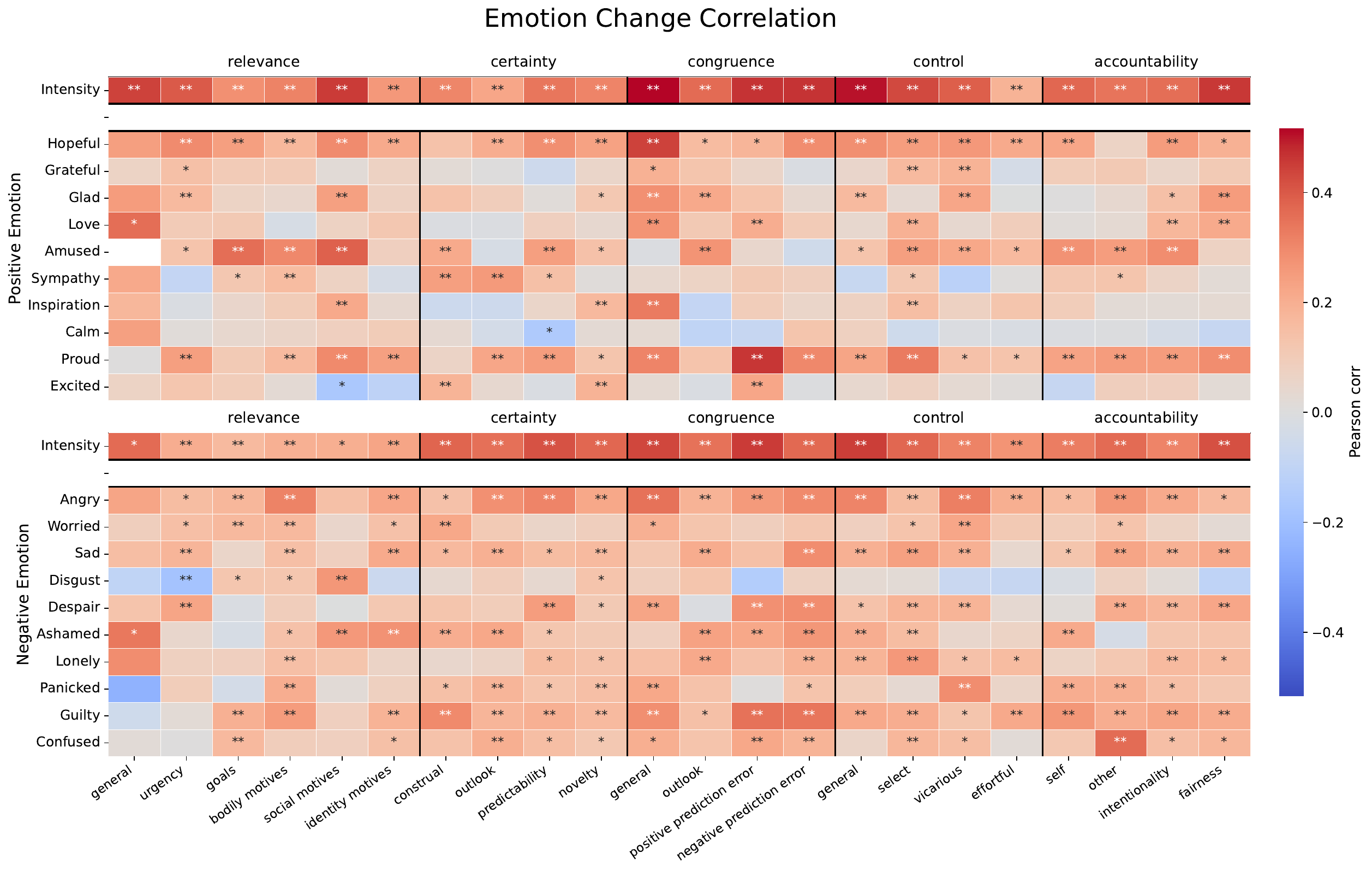}
    \caption{Pearson correlation between model-predicted emotion change and human annotation change after substituting first-person appraisal reasoning with third-person reasoning (Qwen). Red indicates positive correlation; $*p < .05$, $**p < .01$.}
    \label{fig:rq3_qwen}
\end{figure*}

\begin{figure*}[ht]
    \centering
    \includegraphics[width=\textwidth]{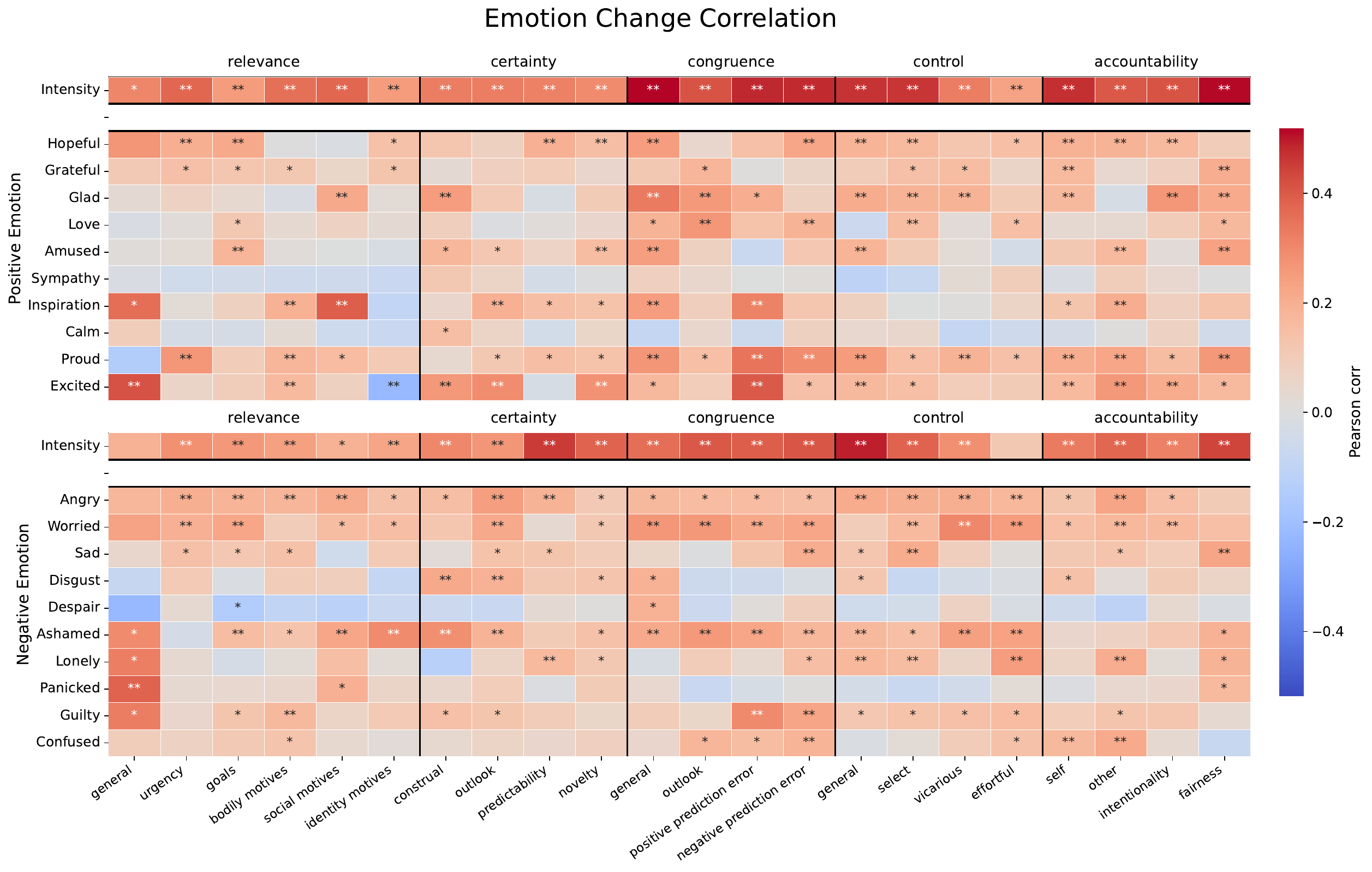}
    \caption{Pearson correlation between model-predicted emotion change and human annotation change after substituting first-person appraisal reasoning with third-person reasoning (Llama). Red indicates positive correlation; $*p < .05$, $**p < .01$.}
    \label{fig:rq3_llama}
\end{figure*}

\begin{figure*}[ht]
    \centering
    \includegraphics[width=\textwidth]{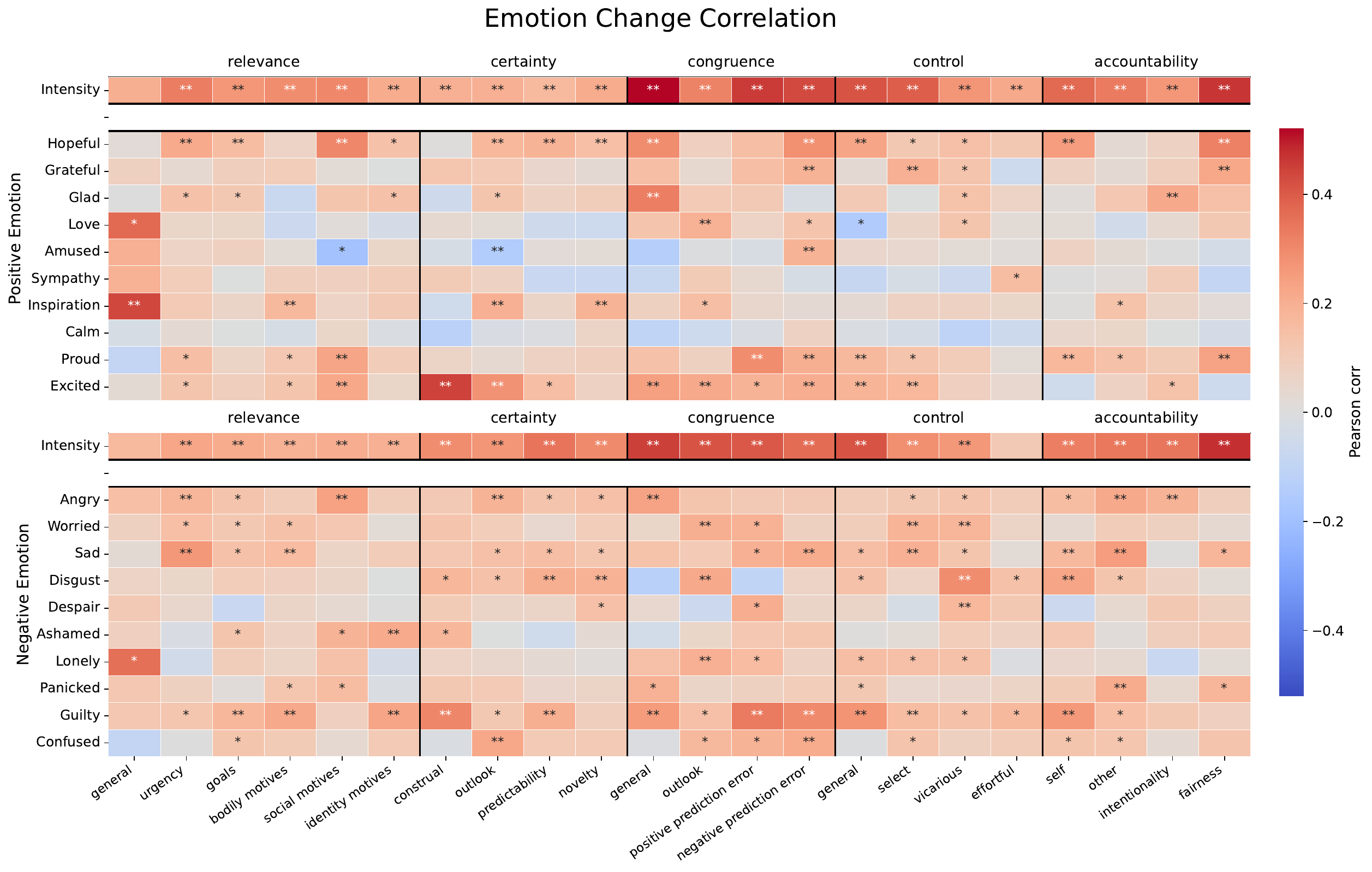}
    \caption{Pearson correlation between model-predicted emotion change and human annotation change after substituting first-person appraisal reasoning with third-person reasoning (PsyLLM). Red indicates positive correlation; $*p < .05$, $**p < .01$.}
    \label{fig:rq3_psyllm}
\end{figure*}

\begin{figure*}[ht]
    \centering
    \includegraphics[width=\textwidth]{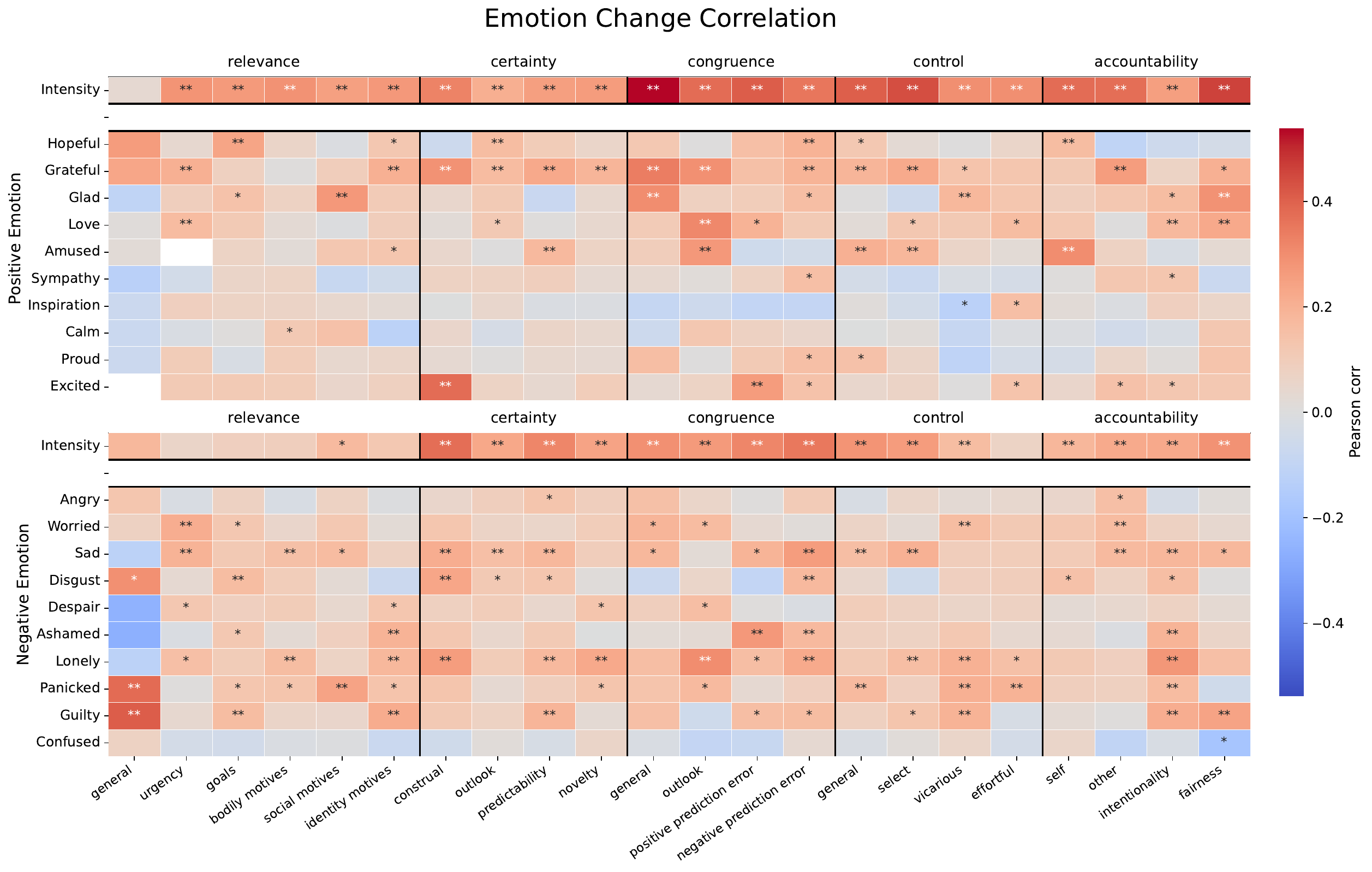}
    \caption{Pearson correlation between model-predicted emotion change and human annotation change after substituting first-person appraisal reasoning with third-person reasoning (EmoLlama). Red indicates positive correlation; $*p < .05$, $**p < .01$.}
    \label{fig:rq3_emollama}
\end{figure*}

Figures~\ref{fig:rq3_claude}-\ref{fig:rq3_emollama} present the Pearson correlation between model-predicted emotion change and human annotation change under appraisal reasoning intervention for each model. Each heatmap shows correlations across all 22 appraisal dimensions (columns) and emotion outputs (rows), covering both emotion intensity and positive and negative emotion labels.

\subsection{Data and Code Access}
\label{app:data-access}

The dataset\footnote{\url{https://huggingface.co/datasets/zhaoyuesun/CAREBecnch}} and code\footnote{\url{https://github.com/ZhaoyueSun/CAREBench/}} are publicly available. The dataset is released under the CC BY-NC-ND 4.0 license, which permits sharing with attribution but prohibits commercial use and derivative works. The code is released under the MIT License.

\subsection{Ethics and Broader Impacts}
\label{app:ethics} 

This study received ethical approval from the institutional review process prior to data collection. Human participants were recruited via Prolific and compensated above the minimum hourly rate; participation was voluntary and all participants provided informed consent prior to data collection. No sensitive personal information beyond demographic attributes was collected.

The primary goal of this work is to advance the scientific understanding of LLMs' emotion understanding capabilities, with potential benefits for applications in affective computing, human-computer interaction, and mental health support. We acknowledge that emotion understanding models could be misused, for instance to infer individuals' emotional states without consent or to manipulate emotional responses in sensitive contexts. To mitigate the risk of misuse, the dataset is released under the CC BY-NC-ND 4.0 license, which prohibits commercial use and derivative works.

\subsection{Licenses for Existing Assets}
\label{app:license}

The licenses for LLMs evaluated in this work are as follows: GPT-5.2 is governed by the OpenAI Terms of Service\footnote{\url{https://openai.com/policies/service-terms/}}; Claude-Sonnet-4.6 is governed by the Anthropic Terms of Service\footnote{\url{https://www.anthropic.com/legal/aup}}; Llama 3.1-8B-Instruct is released under the Llama 3.1 Community License\footnote{\url{https://www.llama.com/llama3_1/license/}}; Qwen 3.5-9B is released under the Apache 2.0 License\footnote{\url{https://huggingface.co/Qwen/Qwen3.5-9B}}; PsyLLM is released under a custom license restricting use to research and educational purposes\footnote{\url{https://huggingface.co/GMLHUHE/PsyLLM-8B}}; and EmoLLaMa-chat-13B is released under the MIT License\footnote{\url{https://huggingface.co/lzw1008/Emollama-chat-13b}}. All models are used in accordance with their respective licenses and terms of service.


\end{document}